%% file: paper_arxiv.tex
\documentclass[sigconf]{acmart} 

\usepackage[utf8]{inputenc}
\usepackage[T1]{fontenc}

\usepackage{graphicx}
\usepackage{amssymb}
\usepackage{comment}
\usepackage{bm}
\usepackage{multirow}
\usepackage{amssymb}
\usepackage{amsmath}
\usepackage{microtype}
\usepackage{listings}
\usepackage{algorithm}
\usepackage{algpseudocode}
\usepackage{mathrsfs}

\setcitestyle{square}

\usepackage{subfigure}
\usepackage{textcomp}

\iffalse
\definecolor{mncolor}{RGB}{255,50,00}
\newcommand\MATTHIAS[1] {\textbf{\textcolor{mncolor}{MN: #1}}}
\definecolor{jtcolor}{RGB}{0,0,255}
\newcommand\JT[1] {\emph{\textcolor{jtcolor}{JT: #1}}}
\definecolor{mzcolor}{RGB}{10,120,10}
\newcommand\MZ[1] {\emph{\textcolor{mzcolor}{MZ: #1}}}
\definecolor{ctcolor}{RGB}{250,100,100}
\newcommand\CT[1] {\emph{\textcolor{ctcolor}{CT: #1}}}
\definecolor{mscolor}{RGB}{255,0,255}
\newcommand\MS[1] {\emph{\textcolor{mscolor}{MS: #1}}}
\newcommand\rev[1] {\emph{\textcolor{mncolor}{#1}}}
\newcommand\TODO[1] {\emph{\textcolor{mncolor}{#1}}}
\else 
\newcommand\MATTHIAS[1] {}
\newcommand\JT[1] {}
\newcommand\MZ[1] {}
\newcommand\CT[1] {}
\newcommand\MS[1] {}
\newcommand\rev[1] {}
\newcommand\TODO[1] {}
\fi

\newcommand{\paragraphbf}[1]{\paragraph{ \textbf{#1}}}

\graphicspath{
{figs/}
}

\acmJournal{TOG}
\acmVolume{36}
\acmNumber{6}
\acmArticle{123}
\acmYear{2017}
\acmMonth{11}

\input{macros}

\begin{document}
\title{FaceVR: Real-Time Gaze-Aware \\ Facial Reenactment in Virtual Reality}

\author{Justus Thies}
\affiliation{%
	\institution{Technical University Munich}}
\email{justus.thies@tum.de}

\author{Michael Zollh{\"o}fer}
\affiliation{%
	\institution{Stanford University}}
\email{zollhoefer@cs.stanford.edu}

\author{Marc Stamminger}
\affiliation{%
	\institution{University of Erlangen-Nuremberg}}
\email{marc.stamminger@fau.de}

\author{Christian Theobalt}
\affiliation{%
	\institution{Max-Planck-Institute for Informatics}}
\email{theobalt@mpi-inf.mpg.de}

\author{Matthias Nie{\ss}ner}
\affiliation{%
	\institution{Technical University Munich}}
\email{niessner@tum.de}


\begin{abstract}
	\input{abstract}
\end{abstract}

\begin{teaserfigure}
   \centering
   \vspace{-0.3cm}
   \includegraphics[width=0.95\textwidth]{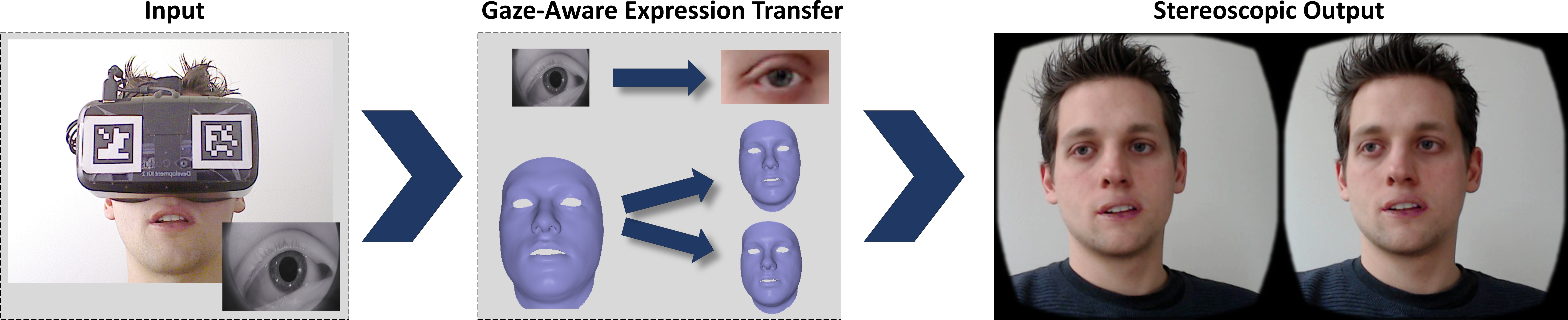}
   \vspace{-0.25cm}
   \caption{We present {\em FaceVR}, a novel method to perform real-time gaze-aware facial reenactment with a virtual reality device (left). In order to capture a face, we use a commodity RGB-D sensor with a frontal view; the eye region is tracked using a new data-driven approach based on data from an IR camera located inside the head-mounted display. Using the 3D reconstructed face as an intermediate, we can modify and edit the face in the target video, as well as re-render it in a photo-realistic fashion, allowing for a variety of applications; e.g., removal of VR goggles or gaze re-targeting. In addition, we render our output in stereo (right), which enables display on stereo devices such as other VR headsets.}
   \label{fig:teaser}
\end{teaserfigure}




%
\setcopyright{none}
\settopmatter{printacmref=false} 
\renewcommand\footnotetextcopyrightpermission[1]{} 
\pagestyle{plain} 

\makeatletter
\def\runningfoot{\def\@runningfoot{}}
\def\firstfoot{\def\@firstfoot{}}
\makeatother

\maketitle

\section{Introduction}
\label{sec:intro}
\input{intro}

\section{Related Work}
\label{sec:related}
\input{related}
\section{Hardware Setup}
\label{sec:setup}
\input{setup}

\section{Synthesis of Facial Imagery}
\label{sec:model}
\input{facemodel}
\section{Parametric Model Fitting}
\label{sec:tracking}
\input{tracking}
\section{An Image-based Eye and Eyelid Model}
\label{sec:learning}
\input{eyemodel}
\section{Face Rig and Compositing}
\label{sec:compositing}
\input{compositing.tex}
\section{Results}
\label{sec:results}
\input{results}

\section{Limitations}
\label{sec:limitations}
\input{limitations}

\section{Conclusion}
\label{sec:conclusion}
\input{conclusion}
\section{Acknowledgments}
We thank Angela Dai for the video voice over and all actors for the VR reenactment.
The facial landmark tracker was kindly provided by TrueVisionSolution.
This research is funded by the German Research Foundation (DFG), grant GRK-1773 Heterogeneous Image Systems, the ERC Starting Grant 335545 CapReal, the Max Planck Center for Visual Computing and Communications (MPC-VCC), a TUM-IAS Rudolf Mößbauer Fellowship, and a Google Faculty Award.

\bibliographystyle{ACM-Reference-Format}
\bibliography{paper}

\include{appendix}

\end{document}

%% file: macros.tex
\DeclareMathOperator*{\argmin}{argmin}

\def \path{\bp C}



%% file: abstract.tex

\nocite{faceVR_ArXiv}

We propose {\em FaceVR}, a novel image-based method that enables video teleconferencing in VR based on self-reenactment.
State-of-the-art face tracking methods in the VR context are focused on the animation of rigged 3d avatars \cite{li2015facial,olszewski2016high}.
While they achieve good tracking performance the results look cartoonish and not real.
In contrast to these model-based approaches, {\em FaceVR} enables VR teleconferencing using an image-based technique that results in nearly photo-realistic outputs.
The key component of FaceVR is a robust algorithm to perform real-time facial motion capture of an actor who is wearing a head-mounted display (HMD), as well as a new data-driven approach for eye tracking from monocular videos.
Based on reenactment of a prerecorded stereo video of the person without the HMD, FaceVR incorporates photo-realistic re-rendering in real time, thus allowing artificial modifications of face and eye appearances.
For instance, we can alter facial expressions or change gaze directions in the prerecorded target video.
In a live setup, we apply these newly-introduced algorithmic components.

%% file: intro.tex

Modern head-mounted virtual reality displays, such as the Oculus Rift\texttrademark~ or the HTC Vive\texttrademark, are able to provide very believable and highly immersive stereo renderings of virtual environments to a user. 
In particular, for teleconferencing scenarios, where two or more people at distant locations meet (virtually) face-to-face in a virtual meeting room, VR displays can provide a far more immersive and connected atmosphere than today's teleconferencing systems.
These teleconferencing systems usually employ one or several video cameras at each end to film the participants, whose video(s) are then shown on one or several standard displays at the other end. 
Imagine one could take this to the next level, and two people in a VR teleconference would each see a photo-realistic 3D rendering of their actual conversational partner, not simply an avatar, but in their own HMD.
The biggest obstacle in making this a reality is that while the HMD allows for very immersive rendering, it is a large physical device which occludes the majority of the face.
In other words, even if each participant of a teleconference was recorded with a 3D video rig, whose feed is streamed to the other end's HMD, natural conversation is not possible due to the display occluding most of the face.
Recent advancements in VR displays are flanked by great progress in face performance capture methods.
State-of-the-art approaches enable dense reconstruction of dynamic face geometry in real-time, from RGB-D~\cite{Weise2011,Bouaziz:2013,Li2013,Zollhoefer2014,Hsieh2015,siegl2017} or even RGB cameras~\cite{Cao2014,Cao2015,thies2016face}. 
A further step has been taken by recent RGB-D~\cite{Thies15} or RGB-only~\cite{thies2016face} real-time facial reenactment methods.
In the aforementioned VR teleconferencing setting, a facial self-reenactment approach can be used to show the unoccluded face of each participant on the VR display at the other end.
Unfortunately, the stability of many real-time face capture methods suffers if the tracked person wears an HMD.
Furthermore, existing reenactment approaches cannot transfer the appearance of eyes, including blinking and eye gaze - yet exact reproduction of the facial expression, including the eye region, is crucial for conversations in VR.

In our work, we therefore propose {\em FaceVR}, a new real-time facial reenactment approach that can transfer facial expressions and realistic eye appearance between a source and a target actor video.
Eye movements are tracked using an infrared camera inside the HMD, in addition to outside-in cameras tracking the unoccluded face regions (see Fig.~\ref{fig:teaser}).
Using the self-reenactment described above, where the target video shows the source actor without the HMD, the proposed approach, for the first time, enables live VR teleconferencing.
In order to achieve this goal, we make several algorithmic contributions:

\begin{itemize}
	\item Robust real-time facial performance capture of a person wearing an HMD, using an outside-in RGB-D camera stream, with rigid and non-rigid degrees of freedom, and an HMD-internal camera.
	\item Real-time eye-gaze tracking with a novel classification approach based on random ferns, for video streams of an HMD-internal camera or a regular webcam.
	\item Facial reenactment with photo-realistic re-rendering of the face region including the mouth and the eyes, using model-based shape, appearance, and lighting capture.
	\item An end-to-end system for facial reenactment in VR, where the source actor is wearing an HMD and the target actor is recorded in stereo.
\end{itemize}

%% file: related.tex

%

A variety of methods exist to capture detailed static and dynamic face geometry with specialized
controlled acquisition setups~\cite{Klehm2015}. 
Some methods use passive multi-view reconstruction in a studio setup~\cite{Borshukov2003,Pighin2006,Beeler2011,Fyffe:2014},
optionally with the support of invisible makeup~\cite{Williams1990} or face markers~\cite{Huang:2011}. 
Methods using active scanners for capture were also developed~\cite{Zhang2004,Weise2009}.
 
Many approaches employ a parametric identity model~\cite{Blanz1999,Blanz2003}, and face expression~\cite{Tena:2011}.
Blend shape models are widely used for representing the expression space~\cite{Pighin1998,Lewis14}, and multi-linear models jointly represent the identity and expression space~\cite{Vlasic2005,Shi2014}.
Newer methods enable dense face performance capture in more general scenes with more lightweight setups, such as a stereo camera~\cite{Valgaerts2012}, or even just a single RGB video at off-line frame rates~\cite{Garrido2013,Suwajanakorn2014,Shi2014,Fyffe:2014}.
Garrido et al.~\shortcite{garrido2016reconstruction} reconstruct a fully controllable parametric face rig including reflectance and fine scale detail, and~\cite{Suwajanakorn_2015_ICCV} build a modifiable mesh model of the face. 
\cite{Ichim2015} reconstruct a game-type 3D face avatar from static multi-view images and a video sequence of face expressions.
More recently, methods reconstructing dense dynamic face geometry in real-time from a single RGB-D camera~\cite{Weise2011,Zollhoefer2014,Bouaziz:2013,Li2013,Hsieh2015} were proposed. 
Some of them estimate appearance and illumination along with geometry~\cite{Thies15}. 
Using trained regressors~\cite{Cao2014,Cao2015}, or parametric model fitting, dense dynamic face geometry can be reconstructed from monocular RGB video~\cite{thies2016face}. 
Recently, Cao et al.~\shortcite{Cao:2016} proposed an image-based representation for dynamic 3D avatars that supports various hairstyles and parts of the upper body.

The ability to reconstruct face models from monocular input data enables advanced image and video editing effects.
Given a portrait of a person, a limitless number of appearances can be synthesized \cite{Kemelmacher-Shlizerman:2016} based on face replacement and internet image search.
Examples for video editing effects are re-arranging a database of video frames~\cite{Li2012} such that mouth motions match a new audio stream~\cite{Bregler1997,TaylorTM15}, face puppetry by reshuffling a database of video frames~\cite{Kemelmacher2010}, or re-rendering of an entire captured face model to make mouth motion match a dubbed audio-track~\cite{Garrido2015}.
Other approaches replace the face identity in a target video~\cite{Dale2011,Garrido2014}. When face expressions are modified, it is often necessary to re-synthesize the mouth and its interior under new or unseen expressions, for which image-based~\cite{Kawai2014,thies2016face} or 3D template-based~\cite{Thies15} methods were examined.
Recently, Suwajanakorn et al.~\shortcite{Suwajanakorn2017} presented a system that learns the mapping between audio and lip motion.
This learning based approach requires a large amount of person specific training data and cannot control the gaze direction.
Vlasic et al.~\shortcite{Vlasic2005} describe a model-based approach for expression mapping onto a target face video, enabling off-line reenactment of faces under controlled recording conditions.
While Thies et al.~\shortcite{Thies15} enable real-time dense tracking and photo-realistic expression mapping between source and target RGB-D video, Face2Face~\cite{thies2016face} enables real-time expression cloning between captured RGB video of one actor and an arbitrary target face video.
Under the hood, they use a real-time tracker capturing dense shape, appearance and lighting. 
Expression mapping and image-based mouth re-rendering enables photo-realistic target appearance.
None of the aforementioned capture and reenactment approaches succeeds under strong face occlusion by a VR headset, nor can combine data from several cameras -- inside and outside the display -- and thus cannot realistically re-render the eye region and appearance, including correct gaze direction. 
Parts of our method are related to image-based eye-gaze estimation approaches. Commercial systems exist for eye gaze tracking of the unoccluded face using special externally placed cameras, e.g., from Tobii\footnote{\url{www.tobii.com}}, or IR cameras placed inside a VR headset, e.g., from Pupil Labs\footnote{\url{www.pupil-labs.com}\label{footnote1}}, FOVE\footnote{\url{www.getfove.com}} or SMI\footnote{\url{www.smivision.com}}.
Appearance-based methods for gaze-detection of the unoccluded face from standard externally placed cameras were also researched~\cite{Sugano2014,Zhang2015}.
Wang et al.~\shortcite{Wang:2016} simultaneously capture 3D eye gaze, head pose, and facial expressions using a single RGB camera at real-time rates.
However, they solve a different problem from ours; we need to reenact -- i.e., photo-realistically synthesize -- the entire eye region appearance in a target video of either a different actor, or the same actor under different illumination, from input video of an in-display camera. 
Parts of our method are related to gaze correction algorithms for teleconferencing where the eyes are re-rendered such that they look into the web-cam, which is typically displaced from the video display~\cite{Criminisi:2003,Kuster12Gaze,Kononenko2015}.
Again, this setting is different from ours, as we need to realistically synthesize arbitrary eye region motions and gazes, and not only correct the gaze direction. 

Related to our paper is the work by Li et al.~\shortcite{li2015facial} who capture moving facial geometry while wearing an HMD with a rigidly attached depth sensor.
In addition, they measure strain signals with electronic sensors to estimate facial expressions of regions hidden by the display.
As a result, they obtain the expression coefficients of the face model which are used to animate virtual avatars.
Recently, Olszewski et al.~\shortcite{olszewski2016high} propose an approach for HMD users to control a digital avatar in real-time based on RGB data.
The user's mouth is captured by a camera that is rigidly attached to the HMD and a convolutional neural network is used to regress from the images to the parameters that control a digital avatar.
They also track eyebrow motion based on a camera that is integrated into the head mounted display.
Both of these approaches only allow to control a virtual avatar -- rather than a real video -- and do not capture the eye motion.
Our approach takes this a step further and captures facial performance as well as the eye motion of a person using an HMD.
In addition, we allow to re-render and reenact the face, mouth, and eye motion of a target stereo stream photo-realistically and in real-time.
Recently, Google presented an approach for HMD removal in the virtual/mixed reality setting \cite{hmd_removal_google}, which shows the great interest in such technology.
Instead of removing the entire HMD, they use translucent rendering techniques to reveal the occluded eye region.
They synthesize the eye region similar to our method~\cite{faceVR_ArXiv}, based on the gaze estimation of an HMD-integrated SMI eye tracker and static face geometry.
In contrast, our approach based on self-reenactment produces a stereo video of the person completely without the HMD.
Furthermore, we present a lightweight eye tracking approach that is able to track eye motions and enables us to synthesize new eye motions in a photo-realistic fashion.

%% file: setup.tex

\begin{figure}
	\centering
	\includegraphics[width=0.9\linewidth]{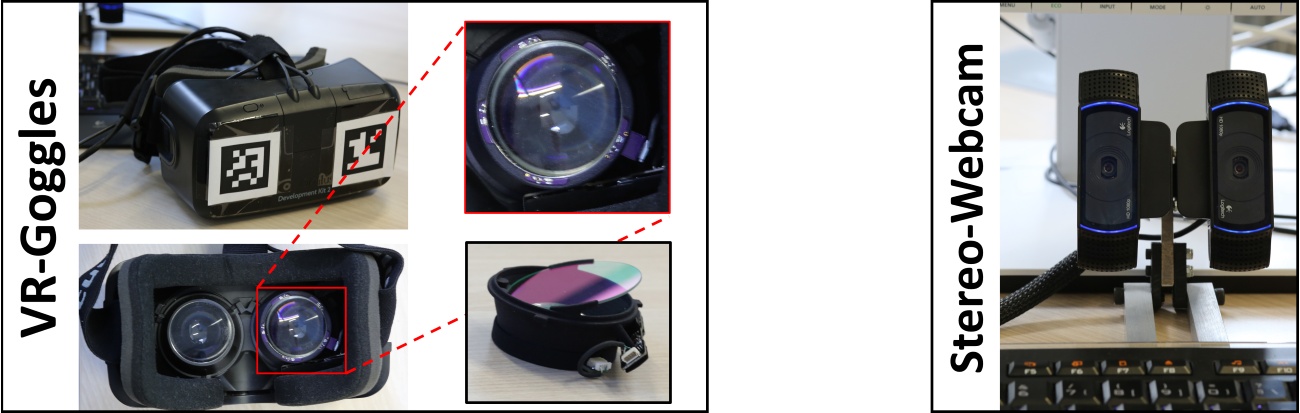}
	\vspace{-0.25cm}
	\caption{Hardware setups: a source actor experiences VR wearing an Oculus DK2 headset (left). We track the source actor using a commodity RGB-D sensor (front-facing), and augment the HMD with ArUco markers, as well as an IR webcam in the inside (mounted with Add-on Cups\textsuperscript{\ref{footnote1}}). The target actor footage is captured with a lightweight stereo rig, which is composed of two webcams (right).}
	\vspace{-0.35cm}
	\label{fig:setup}
\end{figure}

Our approach requires two different inputs.
One is called {\em source}, it is the live video feed of the person wearing a head-mounted display (HMD).
We call the person in this video {\em source actor}.
In addition to this live video, we require a prerecorded stereo video of the person without the HMD.
This stereo video is the {\em target video} and the person in that video is called {\em target actor}.
Note that for self-reenactment the {\em source} and {\em target actor} are the same person.
The {\em source actor} is wearing a head-mounted display (HMD), and we use a lightweight hardware setup to reconstruct and track the source actor's face.
To this end, we augment commodity VR goggles with a simple IR webcam on the inside for tracking one eye.
For tracking the rigid pose and facial expressions, we use outside-in tracking based on a real-time RGB-D sensor (Asus Xtion Pro), as well as ArUco AR markers on the front panel of the HMD.
The tracking and reconstruction pipeline for the {\em target actor} differs. 
Here, we use a stereo setup which is composed of two commodity webcams.
This allows for robust face tracking and generation of 3D video content that we can display on an HMD (which is the case for VR teleconferencing).
We pre-record the target actor's video stream, but we modify and replay it in real-time.
In addition, we assume that the face in the target video is mostly unoccluded.

\subsection{Head-Mounted Display for the Source Actor}\label{sec:setup_source}

To enable VR teleconferencing, we use an Oculus Rift DK2 head-mounted display, and we integrate a simple IR webcam to track the source actor's eyes.
The camera is integrated inside the HMD with Oculus Rift DK2 Monocular Add-on Cups, which allows us to obtain a close-up camera stream of the right eye~\cite{pupilLabs}; see Fig.~\ref{fig:setup}, left.
Although we present results on this specific setup, our method is agnostic to the head-mounted display, and can be used in combination with any other VR device, such as the VR Box, Samsung Gear VR, or HTC Vive.
The monocular camera, which we integrate in the DK2, captures an IR stream of the eye region at a resolution of $640\times480$ pixels at $120$Hz.
IR LEDs are used as active light sources such that bright images can be obtained, and the camera latency is $5.7$ms.
The camera is mounted on the top of the VR device lens and an IR mirror is used to get a frontal view of the eye without interfering with the view on the display.
The camera is located close to the lenses (see Fig.~\ref{fig:setup}, left), and captures images $\mathbf{I}_{\mathcal{E}}$ of the eye at real-time rates.
Note that our prototype has only one internal camera.
Thus, we use the stream of the right eye to infer and reenact the motion of both the left and the right eye.
This is feasible as long as we can assume that the focus distance is the same as during calibration, that is eye vergence (squinting) does not change.
If this assumption does not hold, a second internal camera for the left eye can be easily integrated into our design.
In addition, we augment the DK2 by attaching two ArUco AR markers to the front of the HMD to robustly track the rigid pose.
During face tracking, this allows us to decouple the rigid head pose from the facial expression parameters by introducing additional soft constraints obtained from the markers.
The combination of marker tracking and joint optimization allows to further stabilize the estimates of the rigid head pose, leading to much higher tracking accuracy (see Fig.~\ref{fig:arucoStabilization}).

\paragraphbf{Tracking of the Source Actor}

For tracking the source actor in real-time, we use a commodity RGB-D camera.
Specifically, we use an Asus Xtion Pro RGB-D sensor that captures RGB-D frames of $640\times480$ pixels at $30$ fps (both color and depth).
In every frame, the camera captures an RGB image $\mathbf{I}_\mathcal{I}$ and a depth image  $\mathbf{D}_\mathcal{I}$, which we assume to be spatially and temporally aligned.
Both images are parameterized by pixel coordinates $\mathbf{p}$, each RGB value is $\mathbf{I}_\mathcal{I} (\mathbf{p}) \in \mathbb{R}^3$.
Depth $\mathbf{D}_\mathcal{I} (\mathbf{p}) \in \mathbb{R}$ is reprojected into the same space as $\mathbf{I}_\mathcal{I}$.
Note that we are only considering visible pixel locations $\mathbf{p} \in \mathcal{P}$ on the face that are not occluded by the HMD.

\subsection{3D Stereo Rig for Target Actor Tracking}\label{sec:setup_target}

In order to obtain a 3D reconstruction of the target actor, we use the binocular image stream of a lightweight stereo rig.
Our setup is composed of two commodity webcams (Logitech HD Pro Webcam C920), which are rigidly mounted side-by-side and facing the same direction on a stereo bar; see Fig.~\ref{fig:setup} (right).
The camera rig captures a stereo stream of two RGB pairs $\mathbf{I}_{\mathcal{I}}^{(c)},~c\in\{1,2\}$ at real-time rates.
The two cameras are synchronized up to $33$ms and capture images at the resolution of $800\times600$ pixels at $30$Hz.
This stereo content is used to capture the target 3D video content.
We calibrate the stereo rig intrinsically and extrinsically using standard OpenCV routines.

%% file: facemodel.tex

We parameterize human heads under general uncontrolled illumination based on a multi-linear face and an analytic illumination model.
A linear PCA basis is used for facial identity \cite{Blanz1999} (geometry and reflectance) and a blendshape basis for the expression variations \cite{Alexander:2009,Cao2014b}.
This results in the spatial embedding of the underlying mesh and the associated per-vertex color information parameterized by linear models, $\mathcal{F}(\mathbf{T}, \boldsymbol{\alpha}, \boldsymbol{\beta}, \boldsymbol{\delta})$ and $\mathcal{C}(\boldsymbol{\beta}, \boldsymbol{\gamma})$, respectively.
The mesh has $106$K faces and $53$K vertices.
Here, $\mathbf{T} \in \mathbb{R}^{4 \times 4}$ models the rigid head pose, $\boldsymbol{\alpha} \in \mathbb{R}^{80}$ the geometric identity, $\boldsymbol{\beta} \in \mathbb{R}^{80}$ the surface reflectance properties, $\boldsymbol{\delta} \in \mathbb{R}^{76}$ the facial expression, and $\boldsymbol{\gamma} \in \mathbb{R}^{3\cdot9}$ the incident illumination situation.
The $3 \times 9$ illumination coefficients encode the RGB illumination based on $9$ Spherical Harmonics (SH) \cite{Ramamoorthi2001} basis functions.
For convenience, we stack all parameters of the model in a vector $\mathcal{X} = (\mathbf{T},\boldsymbol{\alpha}, \boldsymbol{\beta}, \boldsymbol{\delta}, \boldsymbol{\gamma}) \in \mathbb{R}^{269}$.
Synthetic monocular images $\mathbf{I}_\mathcal{S}$ and synthetic stereo pairs $(\mathbf{I}_\mathcal{S}^{(1)},\mathbf{I}_\mathcal{S}^{(2)})$ of arbitrary virtual heads can be generated by varying the parameters $\mathcal{X}$ and using the GPU rasterization pipeline to simulate the image formation process.
To this end, we use a standard pinhole camera model $\Pi \left(\bullet \right)$ under a full perspective projection.

\paragraphbf{Mouth Interior} The parametric head model does not contain rigged teeth, a tongue or a mouth interior, since these facial features are challenging to reconstruct and track from stereo input due to strong occlusions in the input sequence.
Instead, we propose two different image-based synthesis approaches (see Sec.~\ref{sec:mouth_interior}).
The first is specifically designed for the self-reenactment scenario, where source and target actor are the same person; here we cross project the mouth interior from the source to the target video.
For arbitrary source and target actor pairs we improved the retrieval strategy of Thies et al.~\shortcite{thies2016face}.
This retrieval approach finds the best suitable mouth frame in a mouth database, captured in a short training sequence.
In contrast to their approach, our retrieval clusters frames into static and dynamic motion segments leading to temporally more coherent results.
The output of this step is then composited with the rendered model using alpha blending (see Sec.~\ref{sec:compositing}).

\paragraphbf{Eyeball and Eyelids}

We use a unified image-based strategy to synthesize plausible animated eyes (eyeball and eyelid) that can be used for photo-realistic facial reenactment in VR applications.
This novel strategy is one of the core components of this work and is described in more detail in Section \ref{sec:learning}.

%% file: tracking.tex

Our approach uses two different tracking and reconstruction pipelines for each (source and target) actor, respectively.
The source actor, who is wearing the HMD, is captured using an RGB-D camera; see Sec.~\ref{sec:setup_source}.
Here, we constrain the face model $\mathcal{F}$ by the visible pixels on the face that are not occluded by the HMD, as well as the attached ArUco AR markers.
The target actor reconstruction -- which becomes the corresponding VR target content that is animated at runtime -- is obtained in a pre-process with the lightweight stereo setup described in Sec.~\ref{sec:setup_target}.
For both tracking pipelines, we use an \textit{analysis-by-synthesis} approach to find the model parameters $\mathcal{X}$ that best explain the input observations.
The underlying inverse rendering problem is tackled based on energy minimization as proposed in \cite{Thies15,thies2016face}.
The tracking for the source and the target actor differ in the energy formulation.
The source actor is partly occluded by the HMD, there we measure dense color and depth alignment based on the observations of the RGB-D camera. 
We restrict the dense reconstruction to the lower part of the face using a predefined visibility mask.
In addition, we use ArUco markers that are attached to the HMD to stabilize the rigid pose of the face (seen 	Fig.~\ref{fig:arucoStabilization}).
As the target videos are recorded in stereo, we adapted the energy formulation to work on binocular RGB data.
The results show that our new stereo tracking approach leads to better tracking accuracy than the monocular tracking of \cite{thies2016face}.
For simplicity, we first describe the energy formulation for tracking the target actor in Sec.~\ref{sec:tracking_target}.
Then, we introduce the objective function for fitting the face model of the source actor in Sec.~ \ref{sec:tracking_source}.

\subsection{Target Actor Energy Formulation}\label{sec:tracking_target}

In order to process the stereo video stream of the target actor, we introduce a model-based stereo reconstruction pipeline that constrains the face model according to both RGB views per frame.
In other words, we aim to find the optimal model parameters $\mathcal{X}$ constrained by the input stereo pair $\{\mathbf{I}_\mathcal{I}^{(c)}\}_{c=1}^{2}$. 
Our model-based stereo reconstruction and tracking energy $E_{\text{target}}$ is a weighted combination of alignment and regularization constraints:
\begin{equation} \label{eq:target}
E_{\text{target}}(\mathcal{X}) = \underbrace{ \Big[ w_{\text{ste}} E_{\text{ste}}(\mathcal{X}) + w_{\text{lan}} E_{\text{lan}}(\mathcal{X})\Big]}_{\text{alignment}} + \underbrace{\Big[ w_{\text{reg}} E_{\text{reg}}(\mathcal{X}) \Big]}_{\text{regularizer}} \enspace{.}
\end{equation}
We use dense photometric stereo alignment $E_{\text{ste}}$ and sparse stereo landmark alignment $E_{\text{lan}}$ in combination with a robust regularization strategy $E_{\text{reg}}$.
The sub-objectives of $E_{\text{target}}$ are scaled based on empirically determined, but constant, weights $w_{\text{ste}}=100$, $w_{\text{lan}}=0.0005$, and $w_{\text{reg}}=0.0025$ that balance the relative importance.

\paragraphbf{Dense Photometric Stereo Alignment}

We enforce dense photometric alignment of the input $\mathbf{I}_\mathcal{I}^{(c)}$ and the synthetic imagery $\mathbf{I}_\mathcal{S}^{(c)}$.
For robustness against outliers, we use the $\ell_{2,1}$-norm \cite{DingZHZ06} instead of a traditional least-squares formulation:
\begin{equation} \label{eq:photo}
E_{\text{ste}}(\mathcal{X}) = \sum_{c=1}^{2}{   \frac{1}{|\mathcal{P}^{(c)}|}\sum_{\mathbf{p} \in \mathcal{P}^{(c)}}{\left\| \mathbf{I}_\mathcal{S}^{(c)}(\mathbf{p}) - \mathbf{I}_\mathcal{I}^{(c)}(\mathbf{p}) \right\|_2} }\enspace{.}
\end{equation}
Here, $\mathcal{P}^{(c)}$ is the set of visible model pixels $\mathbf{p}$ from the $c^{th}$-camera.
The visible pixels of the model are determined by a forward rendering pass using the old parameters.
We normalize based on the total number of pixels $|\mathcal{P}^{(c)}|$ to guarantee that both views have the same influence.
Note that the two sets of visible pixels are updated in every optimization step, and for the forward rendering pass we use the face parameters of the previous iteration or frame.

\paragraphbf{Sparse Stereo Landmark Alignment}

We use sparse point-to-point alignment constraints in 2D image space that are based on per-camera sets $\mathcal{L}^{(c)}$ of $66$ automatically detected facial landmarks. 
The landmarks are obtained by a commercial implementation\footnote{TrueVisionSolutions Pty Ltd} of the detector of Saragih et al.~\shortcite{Saragih2011}:
\begin{equation} \label{eq:feature}
E_{\text{lan}}(\mathcal{X}) = \sum_{c=1}^{2}{ \frac{1}{|\mathcal{L}^{(c)}|}\sum_{(\mathbf{l}, k) \in \mathcal{L}^{(c)}} {w_{\mathbf{l},k} \left\| \mathbf{l} - \Pi( \mathcal{F}_k(\mathbf{T}, \boldsymbol{\alpha}, \boldsymbol{\beta}, \boldsymbol{\delta}) \right\|^2_2} } \enspace{.}
\end{equation}
The projected vertices $\mathcal{F}_k(\mathbf{T}, \boldsymbol{\alpha}, \boldsymbol{\beta}, \boldsymbol{\delta})$ are enforced to be spatially close to the corresponding detected 2D feature $\mathbf{l}$.
Constraints are weighted by the confidence measures $w_{\mathbf{l},k}$, which are provided by the sparse facial landmark detector.

\paragraphbf{Statistical Regularization}

In order to avoid implausible face fits, we apply a statistical regularizer to the unknowns of $\mathcal{X}$ that are based on our parametric face model.
We favor plausible faces where parameters are close to the mean with respect to their standard deviations $\sigma_{\text{id}}$, $\sigma_{\text{alb}}$, and $\sigma_{\text{exp}}$.
\begin{equation} \label{eq:regularizer}
E_{\text{reg}}(\mathcal{X}) = 
\sum_{i=1}^{80} {\left[
	\left(\frac{\boldsymbol{\alpha}_i}{\sigma_{\text{id},i}}\right)^2 +
	\left(\frac{\boldsymbol{\beta}_i}{\sigma_{\text{alb},i}}\right)^2
	\right]} +
\sum_{i=1}^{76} {
	\left(\frac{\boldsymbol{\delta}_i}{\sigma_{\text{exp},i}}\right)^2
} \enspace{.}
\end{equation}
$\sigma_{\text{id}}$ and $\sigma_{\text{alb}}$ are the standard deviations of the statistical face model, $\sigma_{\text{exp}}$ is set to a constant value ($=1$).

\subsection{Source Actor Tracking Objective}\label{sec:tracking_source}

\begin{figure}
	\centering
	\includegraphics[width=0.9\linewidth]{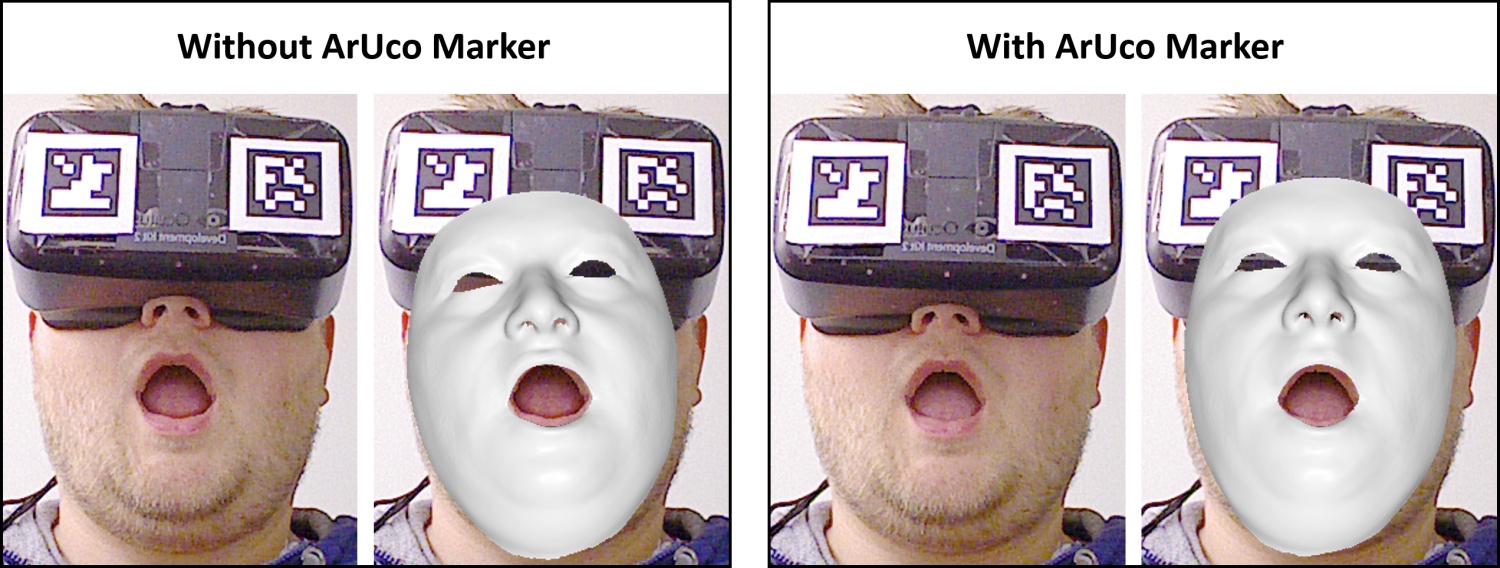}
	\vspace{-0.3cm}
	\caption{Tracking with and without ArUco Marker stabilization.}
	\vspace{-0.3cm}
	\label{fig:arucoStabilization}
\end{figure}

At runtime, we track the source actor who is wearing the HMD and is captured by the RGB-D sensor.
The tracking objective for visible pixels that are not occluded by the HMD is similar to the symmetric point-to-plane tracking energy in Thies et al.~\shortcite{Thies15}.
In addition to this, we introduce rigid stabilization constraints which are given by the ArUco AR markers in front of the VR headset.
These constraints are crucial to robustly separate the rigid head motion from the face identity and pose parameters (see Fig.~\ref{fig:arucoStabilization}).
The total energy for tracking the source actor at runtime is given by the following linear combination of residual terms:
\begin{equation} \label{eq:energy_hmd}
\footnotesize
E_{\text{source}}(\mathcal{X}) = w_{\text{rgb}} E_{\text{rgb}}(\mathcal{X}) + w_{\text{geo}} E_{\text{geo}}(\mathcal{X}) + w_{\text{sta}} E_{\text{sta}}(\mathcal{X}) + w_{\text{reg}} E_{\text{reg}}(\mathcal{X}) \enspace{.}
\end{equation}
The first term of this objective $E_{\text{rgb}}$ measures the photometric alignment of the input RGB image $\mathbf{I}_\mathcal{I}$ from the camera and the synthetically-generated rendering $\mathbf{I}_\mathcal{S}$:
\begin{equation} \label{eq:energy_rgb}
E_{\text{rgb}}(\mathcal{X}) =  \frac{1}{|\mathcal{P}|}\sum_{\mathbf{p} \in \mathcal{P}}{\left\| \mathbf{I}_\mathcal{S}(\mathbf{p}) - \mathbf{I}_\mathcal{I}(\mathbf{p}) \right\|_2} \enspace{.}
\end{equation}
This color term is defined over all visible pixels $\mathcal{P}$ in the bottom half of the face that are not occluded by the HMD, and we use the same $\ell_{2,1}$-norm as in Eq.~\ref{eq:photo}.

In addition to the photometric alignment, we constrain the face model by the captured range data:
\begin{equation} \label{eq:energy_geo}
E_{\text{geo}}(\mathcal{X}) = w_{\text{point}} E_{\text{point}}(\mathcal{X}) + w_{\text{plane}} E_{\text{plane}}(\mathcal{X}) \enspace{.}
\end{equation}
Similar to $E_{\text{rgb}}$, geometric residuals of $E_{\text{geo}}$ are defined over the same set of visible pixels on the face.
The geometric term is composed of two sub-terms, a point-to-point $E_{\text{point}}$ term, where $\mathbf{D}_\mathcal{I}$ is the input depth and $\mathbf{D}_\mathcal{S}$ is the rendered depth (both are back-projected into camera space),
\begin{equation} \label{eq:energy_point}
E_{\text{point}}(\mathcal{X}) = \sum_{\mathbf{p} \in \mathcal{P}}{\left\| \mathbf{D}_\mathcal{S}(\mathbf{p}) - \mathbf{D}_\mathcal{I}(\mathbf{p}) \right\|^2_2} \enspace{,}
\end{equation}
as well as a symmetric point-to-plane term
\begin{equation}
E_{\text{plane}}(\mathcal{X}) = \sum_{\mathbf{p} \in \mathcal{P}}{ \left[ d^2_{\text{plane}}(N_\mathcal{S}(\mathbf{p}), \mathbf{p})  + d^2_{\text{plane}}(N_\mathcal{I}(\mathbf{p}), \mathbf{p}) \right] }, 
\end{equation}
where $d_{\text{plane}}(\mathbf{n}, \mathbf{p}) = \left[ ( \mathbf{D}_\mathcal{S}(\mathbf{p}) - \mathbf{D}_\mathcal{I}(\mathbf{p}) )^T \cdot \mathbf{n} \right]$, $N_\mathcal{I}(\mathbf{p})$ is the input normal and $N_\mathcal{S}(\mathbf{p})$ the rendered model normal.

In addition to the constraints given by the raw RGB-D sensor data, the total energy of the source actor $E_{\text{source}}$ incorporates rigid head pose stabilization.
This is required, since in our VR scenario the upper part of the face is occluded by the HMD.
Thus, only the lower part can be tracked and the constraints on the upper part of the face, which normally stabilize the head pose, are missing.
To stabilize the rigid head pose, we use the two ArUco markers that are attached to the front of the HMD (see Fig.~\ref{fig:arucoStabilization}).

We first extract a set of eight landmark locations based on the two markers (four landmarks each).
In order to handle noisy depth input, we fit two 3D planes to the frame's point cloud that bound each marker, respectively.
We then use the resulting 3D corner positions of the markers, and project them into face model space.
Using these stored reference positions $A_k$ we establish the rigid head stabilization energy $E_{\text{sta}}$:
\begin{equation} \label{eq:aruco}
E_{\text{sta}}(\mathcal{X}) = \frac{1}{|\mathcal{S}|}\sum_{(\mathbf{l}, k) \in \mathcal{S}} {\left\| \mathbf{l} - \Pi( \mathbf{T} A_k) \right\|^2_2} \enspace{.}
\end{equation}
%
Here, $\mathcal{S}$ defines the correspondences between the detected 2D landmark positions $\mathbf{l}$ in the current frame and the reference positions $A_k$.
In contrast to the other data terms, $E_{\text{sta}}$ depends only on the rigid transformation $\mathbf{T}$ of the face and replaces the facial landmark term used by Thies et al.~\shortcite{Thies15}.
Note that the Saragih tracker is unable to robustly track landmarks in this scenario since only the lower part of the face is visible.
The statistical regularization term $E_{\text{reg}}$ is the same as for the target actor (see Eq.~\ref{eq:regularizer}).

\subsection{Data-Parallel Optimization}

We find the optimum of both face tracking objectives $\mathcal{X}_{\text{source}}^* = \argmin_{\mathcal{X}}{E_{\text{source}}(\mathcal{X})}$ and $\mathcal{X}_{\text{target}}^* = \argmin_{\mathcal{X}}{E_{\text{target}}(\mathcal{X})}$ based on variational energy minimization, leading to an unconstrained non-linear optimization problem.
Due to the robust $\ell_{2,1}$-norm used to enforce photo-metric alignment, we find the minimum based on a data-parallel \textit{Iteratively Re-weighted Least Squares} (IRLS) solver \cite{thies2016face}.
At the heart of the IRLS solver, a sequence of non-linear least squares problems are solved with a GPU-based \textit{Gauss-Newton} approach \cite{Zollhoefer2014,wu2014sfs,zollhoefer2015shading,Thies15,devito2016opt,thies2016face} that builds on an iterative Preconditioned Conjugate Gradient (PCG) solver.
The optimization is run in a coarse-to-fine fashion using a hierarchy with three levels.
We only run tracking on the two coarser levels using seven IRLS steps on the coarsest and one on the medium level.
For each IRLS iteration, we perform one GN step with four PCG steps.
In order to exploit temporal coherence, we initialize the face model with the optimization results from the previous frame.
First, this gives us a good estimate of the visible pixel count in the forward rendering pass, and second, it provides a good starting point for the GN optimization.
Note that we never explicitly store $J^T J$, but instead apply the multiplication of the Jacobian (and its transpose) on-the-fly within every PCG step.
Thus, the compute cost for each PCG iteration becomes more expensive for multi-view optimization of $E_{\text{target}}$; although materialization is still less efficient, since we only need a small number of PCG iterations.

%% file: eyemodel.tex

We propose a novel image-based retrieval approach to track and synthesize the region of the eyes, including eyeballs and eyelids.
This approach is later used in all presented applications,especially in the self-reenactment for video conferencing in VR (see Sec.~\ref{sec:gaze_selfreenactment}).
We chose an image-based strategy, since it is specific to a person; it not only models the behavior of the eyeballs, but also captures idiosyncrasies of eyelid movement while enabling photo-realistic re-rendering.
Our approach uses a hierarchical variant of random ferns \cite{Ozuysal:2010} to robustly track the eye region.
To this end, we propose a novel actor-specific and fully automatic training stage.
In the following, we describe our fully automatic data generation process, the used classifier and the optimizations that are required to achieve fast, robust, and temporally stable gaze estimates.

\subsection{Training Data Generation}
\label{sec:learning:database}

To train our image-based eye regression strategy, we require a sufficiently large set of labeled training data.
Since manual data annotation for every new user is practically infeasible, we propose a very efficient approach based on a short eye calibration sequence.

\begin{figure}
	\centering
	\includegraphics[width=0.95\linewidth]{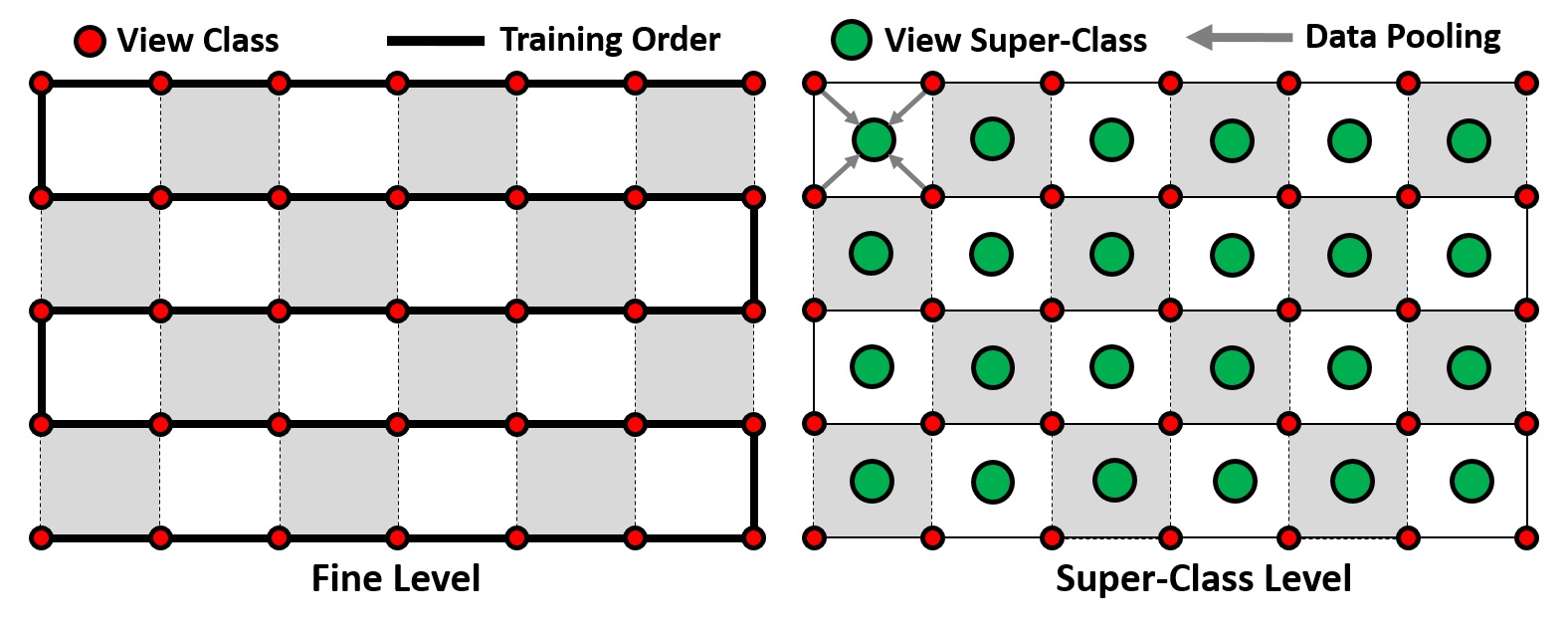}
	\vspace{-0.5cm}
	\caption{Left: the eye calibration pattern used to generate training data for learning our image-based eye-gaze retrieval. In the training phase, we progress row-by-row in a zig-zag order; each grid point is associated with an eye-gaze direction. Right: to obtain robust results, we perform a hierarchical classification where classes of the finer level are accumulated into a smaller set of super classes.}
	\vspace{-0.3cm}
	\label{fig:eyeCalibPattern}
\end{figure}

During the training process, we display a small circle at different positions of a $7\times5$-tiled image grid on the screen in front of the user; see Fig.~\ref{fig:eyeCalibPattern}, left.
This allows us to capture the space of all possible look-at points on the display.
In addition, we capture an image of a closed eye for the synthesis of eye blinks.
The captured image data $\mathcal{I}_n$ is divided into $36 = 7\times5+1$ unique classes $l_n$, where every class is associated with a view direction.
The ground truth gaze directions are given by the current position of the {\em dot} on the screen in the training data.
During training, the user focuses on the displayed dot with his eye gaze.
We show every dot for 2 seconds for each location. 
The data captured in the first $0.4$ seconds is rejected to allow the user a grace period to adjust his eye-gaze to new positions.
In the remaining $1.6$ seconds, we capture $50$ frames which we use to populate the corresponding class.
After that, we proceed to the next class, and move the dot to the next position.
Note that the dot location for a given class is fixed, but we obtain multiple samples within each class (one for each frame) from the input data.
This procedure progresses row-by-row in a zig-zag order; see Fig.~\ref{fig:eyeCalibPattern}, left.
Finally, we augment the samples in each class by jittering each captured source image by $\pm 1$ pixels, resulting in $9 \times 50$ training frames per class.
Each cluster is also associated with a representative image of the eye region obtained from the captured input data.
The representative image of each class is given by the median of the corresponding video clip, which is later used for the synthesis of new eye movements.
Finally, we add an additional class which represents eye blinks; this class is obtained by asking the user to close his eyes at the end of the training phase.
This calibration sequence is performed for both the source and target actor.
Since the calibration sequence is the same for both actors, we obtain one-to-one correspondences between matching classes across actors.
Note, for the source actor we directly use the image data that we observe from the IR camera that is integrated into the HMD as training data.
For the target actor, we compute a normalized view of the eye from the stereo video input using the texture space of the parametric face model.
These normalized views are later used to re-synthesize the eye motions of the target actor (see Sec.~\ref{sec:compositing}).
As detailed in the following subsections, we use the data of the source actor to train an eye-gaze classifier which predicts gaze directions for the source actor at runtime.
Once trained, for a given source input frame, the classifier identifies cluster representatives from the target actor eye data.
The ability to robustly track the eye direction of the source actors forms the basis for real-time gaze-aware facial reenactment; i.e., we are able to photo-realistically animate/modify the eyes of a target actor based on a captured video stream of the source actor.
In the following, we detail our eye tracking strategy.

\subsection{Random Ferns for Eye-gaze Classification}
\label{sec:gaze_classification}

The training data $\{\mathcal{I}_n, l_n\}_{n=1}^{N}$, which is obtained as described in the previous section, is a set of $N$ input images $\mathcal{I}_{n}$ with associated class labels $l_n$.
Each label $l_n \in \{c_l\}_{l=1}^{C}$ belongs to one of $C$ classes $c_l$.
In our case, the images of the eye region are clustered based on gaze direction.
We tackle the associated supervised learning problem by an ensemble of $M$ random ferns \cite{Ozuysal:2010}, where each fern is based on $S$ features.
To this end, we define a sequence of $K=MS$ binary intensity features $\mathbf{F}=\{f_k\}_{k=1}^{K}$, which is split into $M$ independent subsets $\mathbf{F}_m$ of size $S$.
Assuming statistical independence and applying \textit{Bayes Rule}, the log-likelihood of the class label posterior can be written as:
\begin{equation}
\log P(c_l | \mathbf{F}) \sim \log \Big[ P(c_l)\cdot\prod_{m=1}^{M}{P(\mathbf{F}_m | c_l)} \Big]~.
\end{equation}
The class likelihoods $P(\mathbf{F}_m | c_l)$ are learned using random ferns.
Each fern performs $S$ binary tests, which discretizes the per-class feature likelihood into $B=2^S$ bins.
At first, we initialize all bins with one to prevent taking the logarithm of zero.
In all experiments, we use $M=800$ ferns with $S=5$ binary tests.
Finally, the class with the highest posterior probability is chosen as the classification result.
Training takes only around $4.9$ms per labeled image, thus training runs in parallel to the calibration sequence. 
Once trained, the best class is obtained in less than $1.4$ms.

\begin{figure}
	\centering
	\includegraphics[width=\linewidth]{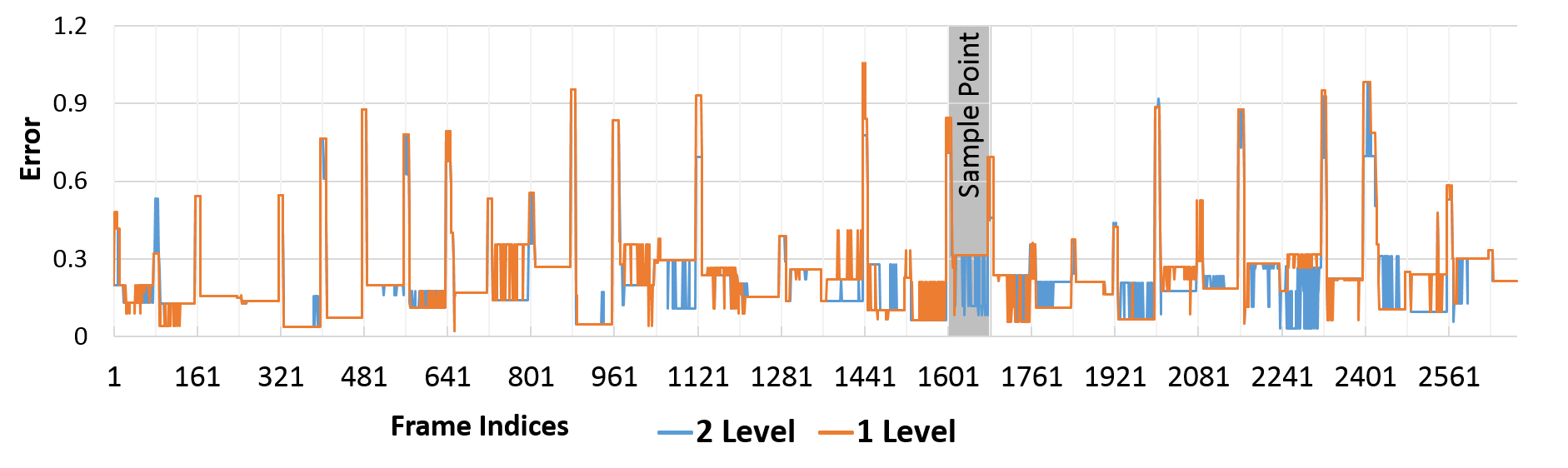}
	\vspace{-0.7cm}
	\caption{Comparison of a one (orange) and a two level (blue) classifier. Ground truth data is obtained by a test subject looking at a dot that appears every $80$ frames ($2.6$ seconds) at random (Sample Point); error is measured in normalized screen space coordinates in $[0,1]^2$. As shown by the magnitude of the positional error, the multi-level classifier obtains higher accuracy.
	}
	\vspace{-0.25cm}
	\label{fig:fernComp}
\end{figure}

\paragraphbf{Hierarchical Eye-gaze Classification}
In order to efficiently handle classification outliers, we perform eye-gaze classification on a two-level hierarchy with a fine and a coarse level.
The $35+1$ classes of the fine level are defined by the grid points of the zig-zag calibration pattern shown in Fig.~\ref{fig:eyeCalibPattern}, left.
To create the coarse level, we merge neighboring classes of the fine level into superclasses.
For a set of four adjacent classes (overlap of one), we obtain one superclass; see Fig.~\ref{fig:eyeCalibPattern}, right.
This leads to a grid with $25=4\times6+1$ unique classes (rather than the $35+1$ classes; the class for eye blink is kept the same).
During training, we train the two hierarchy levels independently. 
The training data for the fine level is directly provided by the calibration pattern, and the data for the coarse level is inferred as described above. 
At test time, we first run the classifier of the coarse level which provides one of the superclasses.
Then the classification on the fine level only considers the four classes of the best matching superclass.
The key insight of this coarse-to-fine classification is to break up the task into easier sub-problems.
That is, the classification on the coarse level is more robust and less prone to outliers of the fern predictions since there are fewer classes to distinguish between.
The fine level then complements the superclass prediction by increasing the accuracy of the inferred eye-gaze directions.
In the end, this multi-level classifier leads to high accuracy results while minimizing the probability of noisy outliers.
In Fig.~\ref{fig:fernComp}, we show a comparison between a one and two level classifier.
The two level approach obtains a lower error (mean $0.217973$, std.dev. $0.168094$) compared to the one level approach (mean $0.24036$, std.dev. $0.18595$).

\paragraphbf{Temporal Stabilization}
We also introduce a temporal stabilizer that favors the previously-retrieved eye-gaze direction.
This particularly helps in the case of small eye motions, where the switch to a new class would introduce unwanted jitter.
To this end, we adjust the likelihood of a specific class $P(c_l)$ using an empirically determined temporal prior such that the previously-predicted eye-gaze direction $c_{\text{old}}$ is approximately $1.05\times$ more likely than changing the state and predicting a different class:
\begin{equation}
P(c_l) =
\begin{cases}
p_1 = \frac{1.05}{1 + 1.05} \approx 0.512,~~~~~\textrm{if } (c_l = c_{\text{old}}) \\
p_2 = 1 - p_1 \, \approx 0.488,~~~~~\textrm{else}
\end{cases}~.
\end{equation}
We integrate the temporal stabilization on both levels of the classification hierarchy.
First, we {\em favor} the super class on the coarse level using the aforementioned temporal prior.
If the current and previous prediction on the coarse level is the same, we apply a similar prior to the view within the superclass.
Otherwise, we use no temporal bias on the fine level.
This allows fast jumps of the eye direction, which is crucial for fast saccade motion that pushes the boundary of the 30Hz temporal resolution of the stereo setup.

%% file: compositing.tex

\begin{figure*}
	\centering
	\includegraphics[width=0.95\linewidth]{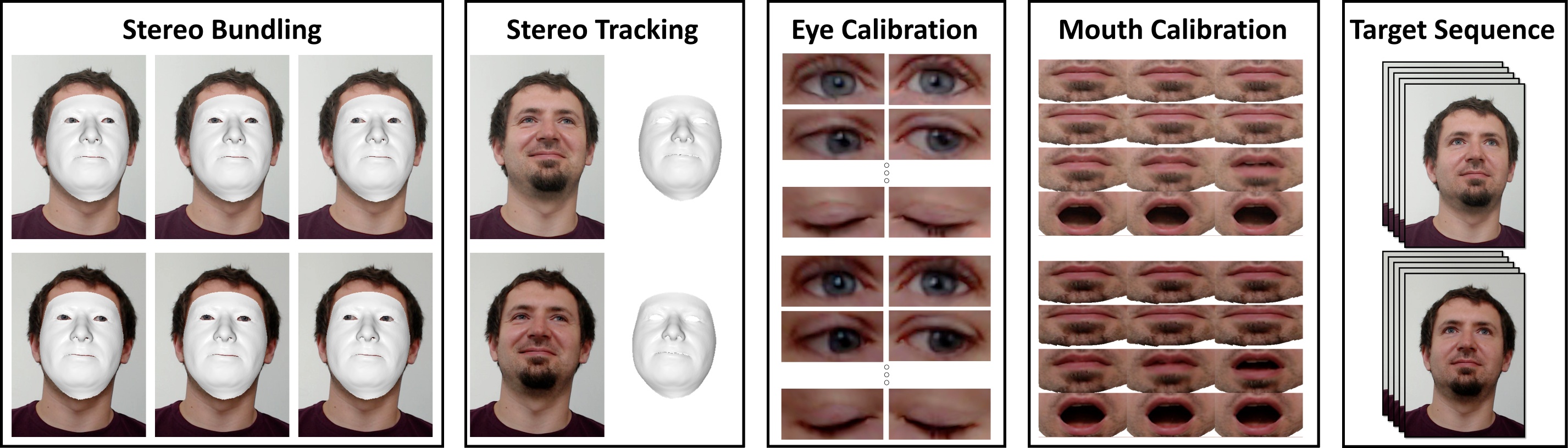}
	\vspace{-0.25cm}
	\caption{Building a personalized stereo avatar; from left to right: we first jointly optimize for all unknowns of our parametric face model using a non-rigid bundle adjustment formulation on the input of three stereo pairs. For tracking, we only optimize for expression, lighting, and rigid pose parameters constrained by synchronized stereo input; this optimization runs in real-time. Next, we train our data-driven eye tracker with data from an eye-calibration sequence. In addition to eye calibration, we build a database of mouth stereo pairs, which captures the variation of mouth motion. Note, the mouth database is only required if the mouth cross-projection is not used. As a result, we obtain a tracked stereo target, which is used during live re-enactment (this is the target actor).}
		\vspace{-0.25cm}
	\label{fig:buildAvatar}
\end{figure*}

\paragraphbf{Generation of a Personalized Face Rig}

At the beginning of each recording, both of the source and target actor, we compute a person-specific face rig in a short initialization stage.
To this end, we capture three keyframes with slightly different head rotations in order to recover the user's facial geometry and skin reflectance.
Given the constraints of these three keyframes, we jointly optimize for all unknowns of our face model $\mathcal{F}$ -- facial geometry, skin reflectance, illumination, and expression parameters -- using our tracking and reconstruction approach.
This initialization requires a few seconds to complete; once computed, we maintain a fixed estimate of the facial geometry and replace the reflectance estimate with a person-specific illumination-corrected texture map.
In the stereo case, we compute one reflectance texture for each of the two cameras.
This ensures that the two re-projections exactly match the input streams, even if the two used cameras have slightly different color response functions.
In the following steps, we use this high-quality stereo albedo map for tracking, and we restrict the optimizer to only compute the per-frame expression and illumination parameters.
All other unknowns (the facial identity) are person-specific and can remain fixed for a given user.
To track and synthesize new eye motions in both videos (source and target), we capture the person-specific appearance and motion of the eyes and eyelids during a short eye-calibration sequence in the initialization stage as described in Sec.~\ref{sec:learning:database}.

\paragraphbf{Reenactment and Real-time Compositing}

At run-time, we use the reconstructed face model along with its calibration data (eye and mouth; see Fig.~\ref{fig:buildAvatar}) to photo-realistically re-render the face of the target actor.
We first modify the facial expression parameters of the reconstructed face model of the target actor to match the face expression of the source actor.
The expressions are transfered from source to target using the subspace deformation transfer approach of Thies et al.~\shortcite{thies2016face}.
In the final compositing stage, we render the mouth texture, the eye textures, and the (potentially modified) 3D face model on top of the target video using alpha blending.
Instead of a static face texture, we use a per-frame texture based on the current frame of the target video.
This leads to results of higher resolution, since slight misalignments during the generation of the personalized face rig have no influence on the final texture quality.

\paragraphbf{Synthesis of Mouth Interior}
\label{sec:mouth_interior}

\begin{figure}
	\centering
	\includegraphics[width=0.8\linewidth]{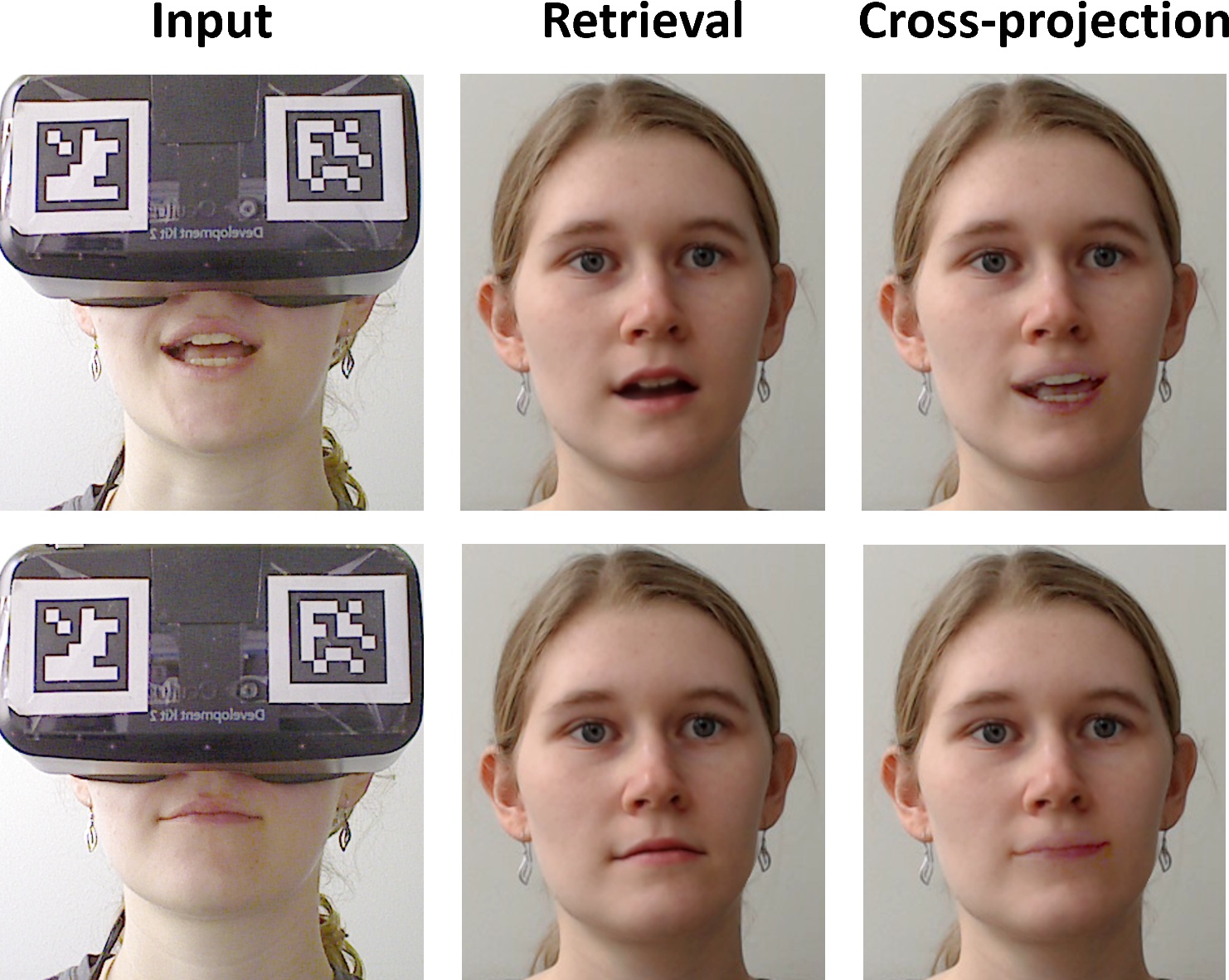}
	\vspace{-0.2cm}
	\caption{Comparison in the case of self-reenactment between the two proposed image-based mouth interior synthesis strategies: Cross-projection (right) leads to more natural and higher quality mouth interiors than the retrieval-based approach.}
	\vspace{-0.25cm}
	\label{fig:poisson_blending_vs_database}
\end{figure}

In order to enable high-quality reenactment of the mouth in the target video, we propose two different approaches.
The method of choice depends on the specific use-case.
In the self-reenactment scenario, which is the case for HMD removal (see Sec.~\ref{sec:gaze_selfreenactment}), we directly project the mouth interior of the source video to the target video.
We use Poisson image blending \cite{Perez:2003} to seamlessly blend the mouth texture into the modified target video.
This ensures an accurate reproduction of the correct mouth shape and interior in the case of identical source and target identity.
The Poisson equation is solved on the GPU using the Jacobi iterative method.
In the case of stereo reenactment, where the source and the target actor differ, we built a database of target mouth interiors using a short calibration sequence as proposed by Thies et al.~\shortcite{thies2016face}.
In this scenario, cross-projection cannot be applied, since this would change the identity of the target actor.
The mouth motion database is clustered into static and dynamic motion segments based on the space-time trajectory of the sparse 2D landmark detections.
We select the mouth frame from the database that has the most similar spatial distribution of 3D marker positions.
In contrast to Thies et al.~\shortcite{thies2016face}, we prefer frames that belong to the same motion segment as the previously retrieved one.
This leads to higher temporal coherence and hence less visual artifacts.
The retrieved mouth frames do not exactly match the transfered facial expression.
To account for this, Thies et al.~\shortcite{thies2016face} stretch the texture based on the face parameterization leading to visual artifacts, i.e., unnaturally stretched teeth, which are temporally unstable.
To alleviate this problem, we propose a new strategy and match the retrieved texture to the outer mouth contour of the target expression using a saliency preserving image warp \cite{Wang:2008}.
For a comparison of both approaches, we refer to the accompanying video.
We use a modified as-rigid-as-possible regularizer that takes local saliency of image pixels into account.
The idea is to deform the mouth texture predominantly in regions that will not lead to visual artifacts.
Stretching is most noticeable for the bright teeth, since they are perfectly rigid in the physical world, while it is harder to detect in the darker regions that correspond to the mouth interior.
Therefore, we use pixel intensity as a proxy to determine local rigidity weights (a high value for bright and low value for dark pixels) that control the amount of warping in different texture regions.
This is based on the assumption that the teeth are the predominant white pixels in the mouth region.
As can be seen in Fig.~\ref{fig:poisson_blending_vs_database}, the mouth cross-projection approach leads to more natural results and captures more details such as the movement of the tongue compared to the retrieval-based approach.

\paragraphbf{Synthesis of the Eye Region}
\label{sec:eye_region}

Our eye gaze estimator is specifically developed to allow a one-to-one correspondence between the source and the target actor (cf.~Sec.~\ref{sec:learning}).
Thus, after tracking the source actor, we know the index of the gaze class in the eye database of the target actor.
To synthesize temporally coherent and plausible eye motion, we temporally filter the eye motion by averaging the retrieved view direction of the gaze class in a small window of frames.
Afterwards, we use the average view direction to perform the texture lookup.
\begin{figure}
	\centering
	\includegraphics[width=1.0\linewidth]{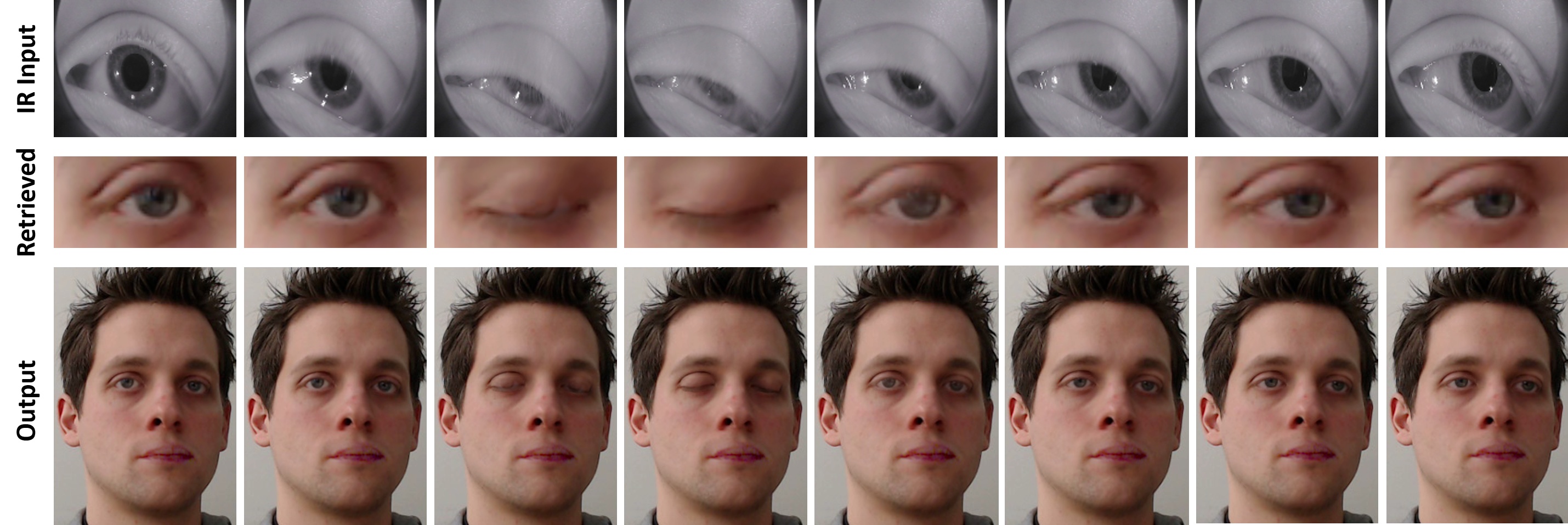}
	\vspace{-0.5cm}
	\caption{Eye Blinking: consecutive frames from left to right. The IR input image captured by the camera mounted inside the HMD (top row) is used to retrieve realistic eye textures (middle row). In the final compositing stage, the texture is seamlessly blended with the target face (bottom row).}
	\vspace{-0.5cm}
	\label{fig:eye_lid_closure}
\end{figure}
As described earlier (Sec.~\ref{sec:gaze_classification}), we use an additional class in our eye gaze classification strategy to represent lid closure.
To obtain temporally smoother transitions between an open and closed eye, we temporally filter the eye texture based on an exponential average (a factor $0.8$ for the retrieved texture and $0.2$ for the last result).
Fig.~\ref{fig:eye_lid_closure} shows an exemplary eye blink transition.
Since the eye images of the target live in the space of the face model texture space, they can directly be used in the final rendering process.

%% file: results.tex

In this section, we evaluate our gaze-aware facial reenactment approach in detail and compare against state-of-the-art tracking methods.
All experiments run on a desktop computer with an Nvidia GTX1080 and a $3.3$GHz Intel Core i7-5820K processor.
For tracking the source and target actor, we use our hardware setup as described in Sec.~\ref{sec:setup}.
Our approach is robust to the specific choice of parameters, and we use a fixed parameter set in all experiments.
For stereo tracking, we set the following weights in our energy formulation: $w_{\text{ste}} = 100.0 $, $w_{\text{lan}} = 0.0005$, $w_{\text{reg}} = 0.0025$. 
Our RGB-D tracking approach uses $w_{\text{rgb} }= 100.0$, $w_{\text{geo}} = 10000.0$, $w_{\text{sta}} = 1.0$, $w_{\text{reg}} = 0.0025$.

As our main result, we demonstrate self-reenactment for VR goggles removal.
In Appendix \ref{sec:appendix} we also show gaze correction in monocular live video footage and gaze-aware facial reenactment.
All three applications share a common initialization stage that is required for the construction of a personalized face and eye/eyelid model of the users; see Sec.~\ref{sec:compositing}.
The source video content is always captured using the Asus Xtion depth sensor.
Depending on the application, we use our lightweight stereo rig or the RGB-D sensor to capture the target actor.

\begin{figure}
	\centering
	\includegraphics[width=0.88\linewidth]{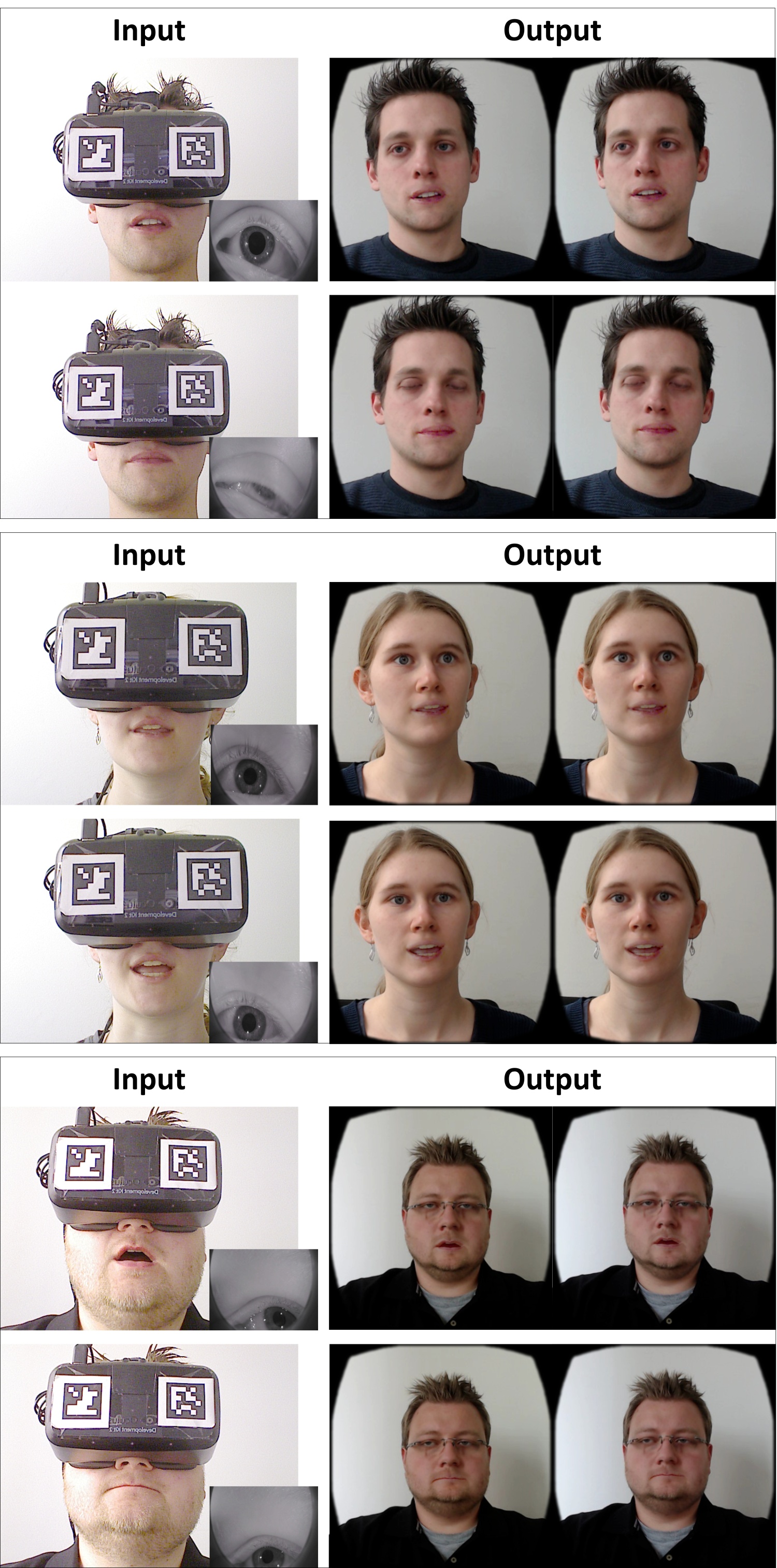}
	\vspace{-0.25cm}
	\caption{Self-Reenactment for VR Video Conferences: our real-time facial reenactment approach allows to virtually remove the HMD by driving a pre-recorded target video of the same person. Note, these results employ the mouth cross-projection strategy to fill in the mouth interior}
	\label{fig:selfreenactment}
\end{figure}

\subsection{Self-Reenactment for VR Video Conferencing} \label{sec:gaze_selfreenactment}

Our real-time facial reenactment approach can be used to facilitate natural video chats in virtual reality.
The major challenge for video conferencing in the VR context is that the majority of the face is occluded by the HMD; therefore, the other person in a VR conversation is unable to see the eye region.
Using self-reenactment, the users can alter both the facial expression and the eye/eyelid motion of the pre-recorded video stream.
This virtually removes the HMD from the face and allows users to appear as themselves in VR without suffering from occlusions due to the head mounted display; see Fig.~\ref{fig:selfreenactment}.
In addition, the output video stream mimics the eye motion, which is crucial since natural eye contact is essential in conversations.
Additionally, we show HMD removal examples with a matching audio stream in the supplemental video.
This shows that, the final result is well aligned with the voice of the source actor.
Although compression is not the main focus of this paper, it is interesting to note that the reenactment results can be easily transferred over a network with low bandwidth.
In order to transmit the 3D video content at runtime to the other participants in a video chat, we only have to send the model parameters, as well as the eye and mouth class indices.
The final modified stereo video can be directly synthesized on the receiver side using our photo-realistic re-rendering.
Given that current video chat software, such as Skype, still struggles under poor network connections, our approach may be able to boost visual quality.
%


\paragraphbf{Evaluation of Face Identity Estimation}
\begin{figure}
	\centering
	\includegraphics[width=0.85\linewidth]{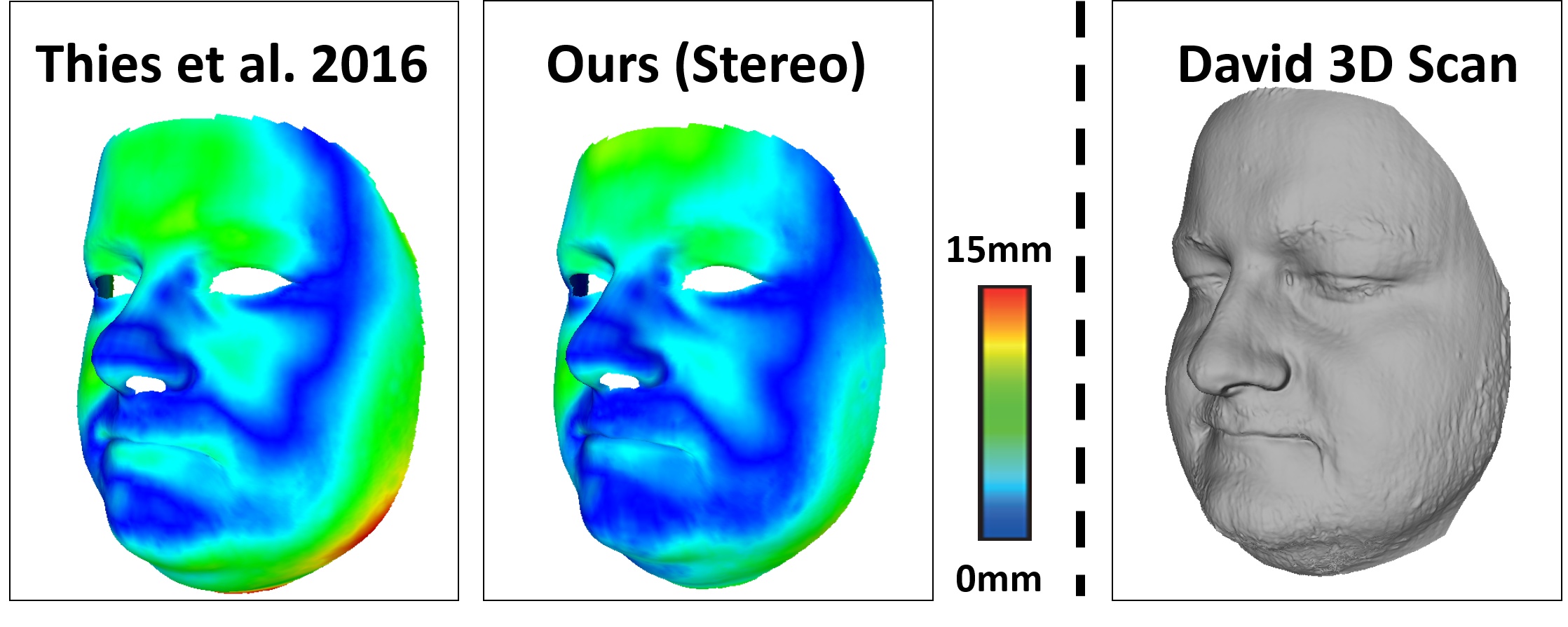}
	\vspace{-0.3cm}
	\caption{Accuracy of reconstructed identity: we compare our result against Face2Face \protect\cite{thies2016face}. Note that our approach obtains a better shape estimate of the chin, nose, and cheek regions.
		For reference, we use a structured light reconstruction from a David 3D scanner.
		The mean Hausdorff Distance of Face2Face is $3.751mm$ (RMSE $4.738mm$). Our approach has a mean distance of $2.672mm$ (RMSE $3.384mm$).
	}
	\vspace{-0.5cm}
	\label{fig:face2faceIdComp}
\end{figure}
The identity of the target actor is obtained using our model-based stereo bundle adjustment strategy.
We compare our identity estimate with the approach of Thies et al.~\shortcite{thies2016face} (Face2Face); see Fig.~\ref{fig:face2faceIdComp}.
As a reference, we use a high-quality structured light scan of the same person taken with a David 3D scanner.
Our approach obtains a better reconstruction of the identity, especially the chin, nose, and cheek regions are of higher quality.
Note that we estimate the identity by model-based bundle adjustment over three stereo pairs.
%
Face2Face uses only the three images of one of the two RGB cameras.

\paragraphbf{Evaluation of Face Tracking Accuracy}
\begin{figure}
	\centering
	\includegraphics[width=0.9\linewidth]{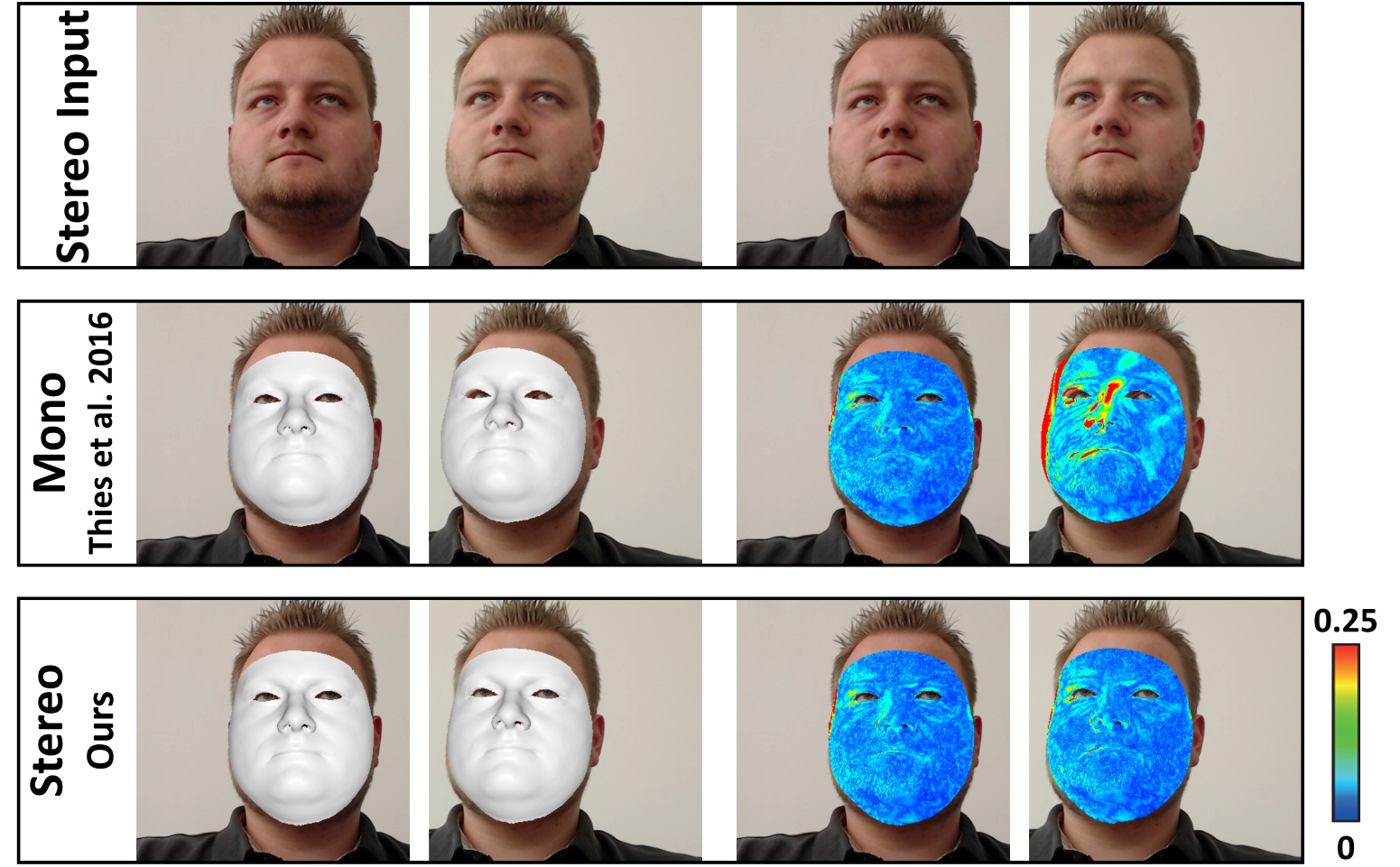}
	\vspace{-0.25cm}
	\caption{Stereo alignment: we compare the photometric alignment accuracy of our approach to Face2Face \protect\shortcite{thies2016face}. Face2Face only obtains a good fit to the image captured by the left camera (average error of $0.011$), but the re-projection to the right camera suffers from strong misalignments (average error of $0.019$). In contrast, our stereo tracking method obtains consistently low errors for both views (average error of $0.011$ left and $0.012$ right).}
	\vspace{-0.2cm}
	\label{fig:face2faceStereoAlignmentComp}
\end{figure}
In Fig.~\ref{fig:face2faceStereoAlignmentComp}, we evaluate the stereo alignment accuracy of our approach and compare to the monocular face tracker of Face2Face \cite{thies2016face}.
As input, we use the binocular image stream captured by our custom stereo setup; see Sec.~\ref{sec:setup}.
We measure the photometric error between the input frames and the re-projection of the tracked face model.
The tracking of Face2Face is based on the left camera stream, since this approach uses only monocular input data.
Thus, Face2Face obtains a good fit with respect to the left camera (average error of $0.011$), but the re-projection regarding the right camera suffers from strong misalignments (average error of $0.019$).
In contrast, our stereo tracking approach obtains consistently low errors for both views (average error of $0.011$ left and $0.012$ right), since we directly optimize for the best stereo overlap.
For the aforementioned re-enactment applications in VR, it is crucial to obtain high-quality alignment with respect to both camera streams of the stereo setup.

\begin{figure}
	\centering
	\includegraphics[width=\linewidth]{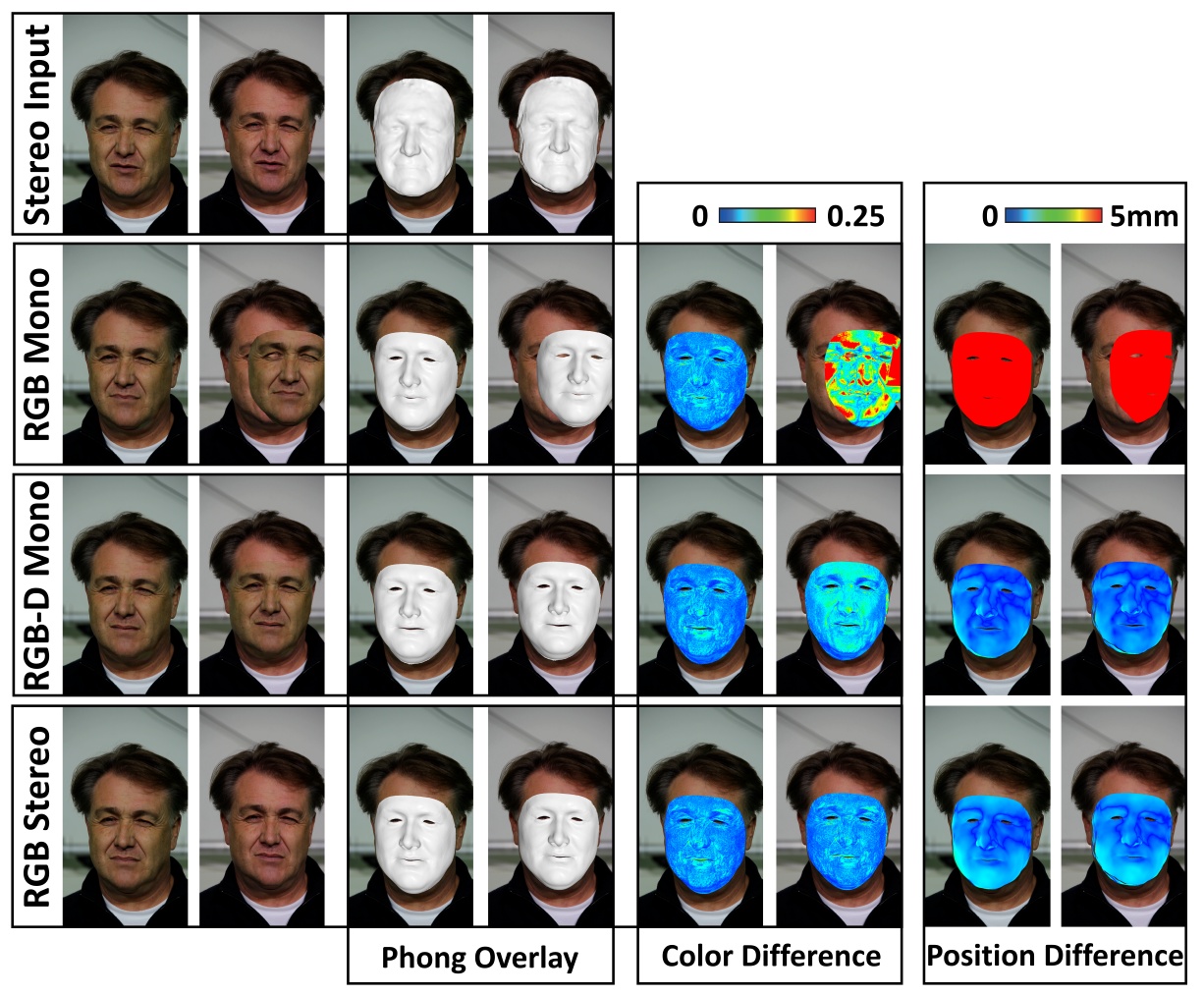}
	\vspace{-0.6cm}
	\caption{Ground truth comparison: we evaluate the photometric and geometric accuracy of our stereo tracking approach (RGB Stereo). As ground truth, we employ the high-quality stereo reconstructions of Valgaerts et al.~\protect\shortcite{Valgaerts2012}. Our approach achieves consistently low photometric and geometric error for both views. We also compare to Thies et al.~\protect\shortcite{Thies15} (RGB-D Mono) and Face2Face \protect\shortcite{thies2016face} (RGB Mono). Both approaches show consistently higher photometric error, since they do not optimize for stereo alignment. Note that the RGB-D tracker uses the ground truth depth as input.}
	\vspace{-0.2cm}
	\label{fig:volker}
\end{figure}

\begin{table}
	\begin{center}
		\begin{tabular}{| l | c | c | c | c | }
			\hline
			& \multicolumn{2}{|c|}{Photometric} & \multicolumn{2}{|c|}{Geometric} \\
			& left & right & left & right \\ \hline
			\multirow{1}{*}{RGB Mono}   & \textbf{0.0130} & 0.0574 & 0.2028 & 0.1994 \\\hline
			\multirow{1}{*}{RGB-D Mono} & \textbf{0.0123} & 0.0183 & \textbf{0.0031} & \textbf{0.0031} \\ \hline
			\multirow{1}{*}{RGB Stereo (Ours)}  & \textbf{0.0118} & \textbf{0.0116} & \textbf{0.0046} & \textbf{0.0046} \\ \hline
		\end{tabular}
		\vspace{0.3cm}
		\caption{Tracking accuracy of our approach (RGB Stereo) compared to Thies et al.~\protect\shortcite{Thies15} (RGB-D Mono) and Face2Face \protect\shortcite{thies2016face} (RGB Mono). Our approach achieves low photometric and geometric errors for both views since we directly optimize for stereo alignment.}
		\vspace{-0.8cm}
		\label{tab:gt_quant}
	\end{center}
\end{table}
We evaluate the accuracy of our approach on ground truth data; see Fig.~\ref{fig:volker}.
As ground truth, we use high-quality stereo reconstructions obtained by Valgaerts et al.~\shortcite{Valgaerts2012}.
To this end, we synthetically generate a high-quality binocular RGB-D stream from the reference data.
Our approach achieves consistently low photometric and geometric errors.
We also compare against the state-of-the-art face trackers of Thies et al.~\shortcite{Thies15} (RGB-D Mono) and Face2Face \cite{thies2016face} (RGB Mono) on the same dataset.
All three approaches are initialized using model-based RGB-(D) bundling of three (stereo) frames.
The RGB Mono and RGB-D Mono trackers show consistently higher photometric errors for the right input stream, since they do not optimize for stereo alignment; see also Tab.~\ref{tab:gt_quant}.
Given that Face2Face \cite{thies2016face} only uses monocular color input, it suffers from depth ambiguity, which results in high geometric errors.
Due to the wrong depth estimate, the re-projection to the right camera image does not correctly fit the input. 
The RGB-D based tracking approach of Thies et al.~\shortcite{Thies15} resolves this ambiguity and therefore obtains the highest depth accuracy.
Note, however, that this approach has access to the ground truth depth data for the sake of this evaluation.
Since the two cameras have slightly different response functions, the reconstructed model colors do not match the right image, leading to high photometric error.
Only our model-based stereo tracker is able to obtain high-accuracy geometric and photometric alignment in both views.
This is crucial for the creation of 3D stereo output for VR applications, as demonstrated earlier.
None of the two other approaches achieves this goal.

\subsection{Evaluation of Eye Tracking Accuracy}

\begin{figure}[h]
	\vspace{-0.4cm}
	\centering
	\includegraphics[width=\linewidth]{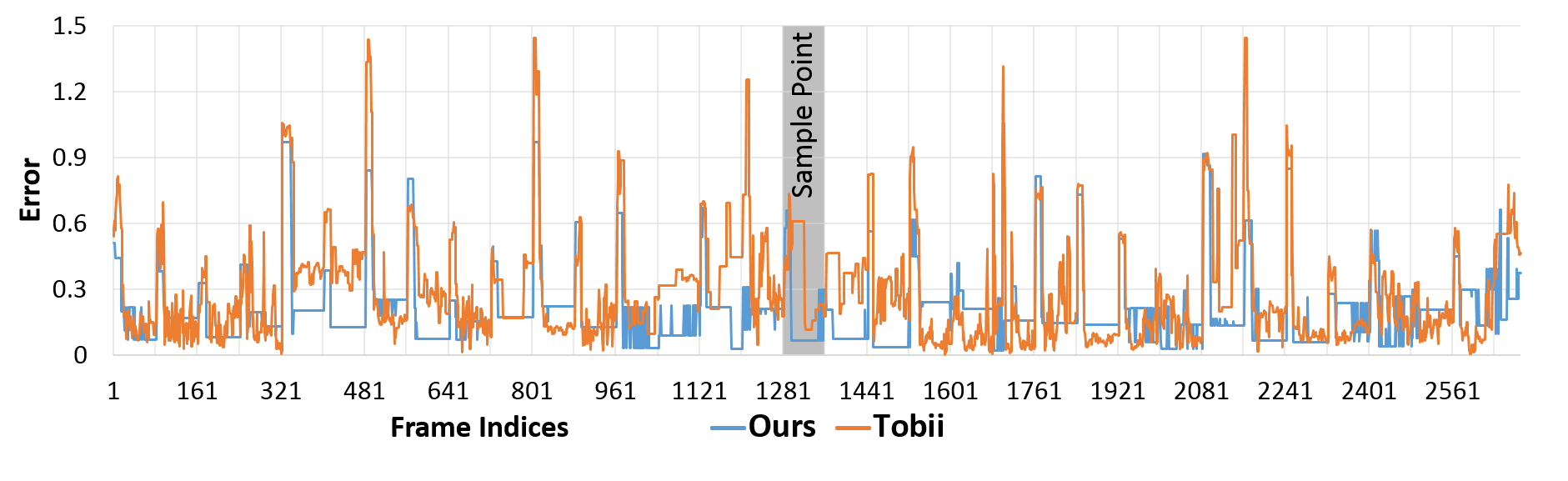}
	\vspace{-0.8cm}
	\caption{Comparison to the commercial Tobii EyeX eye tracking solution. The ground truth data is obtained by a test subject looking at a dot on the screen that appears every $80$ frames ($2.6$ seconds) at random (Sample Point); error is measured in normalized screen space coordinates in $[0,1]^2$. We plot the magnitude of the positional error of Tobii EyeX (orange) and our approach (blue). Our approach obtains a consistently lower error.}
	\vspace{-0.2cm}
	\label{fig:tobiiComp}
\end{figure}

We evaluate the accuracy of our monocular eye gaze classification strategy on ground truth data and compare to the commercial Tobii EyeX eye tracker\footnote{\url{www.tobii.com/xperience/}}.
To this end, a test subject looks at a video sequence of a dot that is displayed at random screen positions for $80$ successive frames ($2.6$ seconds given $30$Hz input) -- this provides a ground truth dataset.
During this test sequence, we capture the eye motion using both the Tobii EyeX tracker and our approach.
We measure the per-frame magnitude of the positional 2D error of Tobii and our approach with respect to the known ground truth screen positions; see Fig.~\ref{fig:tobiiComp}.
Note that screen positions are normalized to $[0,1]^2$ before comparison.
As can be seen, we obtain consistently lower errors.
On the complete test sequence (more than $74$ seconds), our approach has a mean error of  $0.206$ (std.~dev.~$0.178$).
In contrast, the Tobii EyeX tracker has a higher error of $0.284$ (std.~dev.~$0.245$).
The high accuracy of our approach is crucial for realistic and convincing eye reenactment results.
Note, the outside-in tracking of Tobii EyeX does not generalize to the VR context, since both eyes are fully occluded by the HMD.
In the supplemental video we also evaluate the influence of head motion on the retrieved eye texture.
As can be seen in the video sequence, the head motion has less impact on the eye texture retrieval.

\begin{figure}
	\centering
	\includegraphics[width=0.9\linewidth]{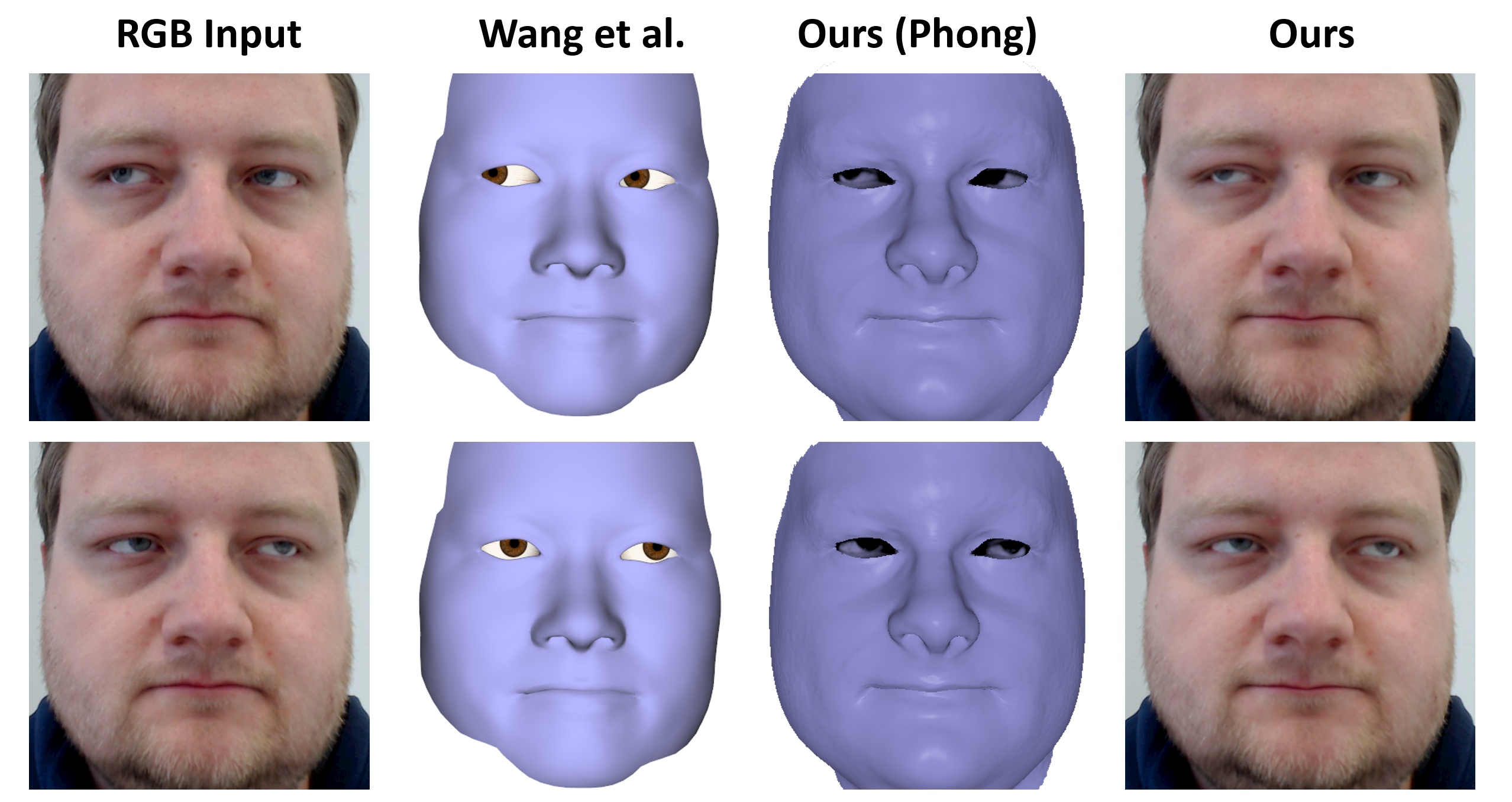}
	\vspace{-0.25cm}
	\caption{Comparison to Wang et al.~\protect\shortcite{Wang:2016}. From left to right: RGB input, output of Wang et al., our phong rendered output with retrieved eyes, final realistic face re-rendering using our approach.}	
	\vspace{-0.2cm}
	\label{fig:comparison_wang}
\end{figure}

We also compare our reconstructions to the state-of-the-art approach of Wang et al.~\shortcite{Wang:2016}, see Fig.~\ref{fig:comparison_wang} (left).
For the complete sequence, we refer to the supplemental video.
Our reconstructions are of similar quality in terms of the obtained facial shape and the retrieved gaze direction.
Note, in contrast to Wang et al.~\shortcite{Wang:2016}, our approach additionally enables realistic re-rendering of the actor, see Fig.~\ref{fig:comparison_wang} (right), which is the foundation for VR goggles removal and reenactment in virtual reality at the cost of a short person specific calibration sequence.

\subsection{Perceptual Evaluation}

\begin{figure}
	\centering
	\includegraphics[width=\linewidth]{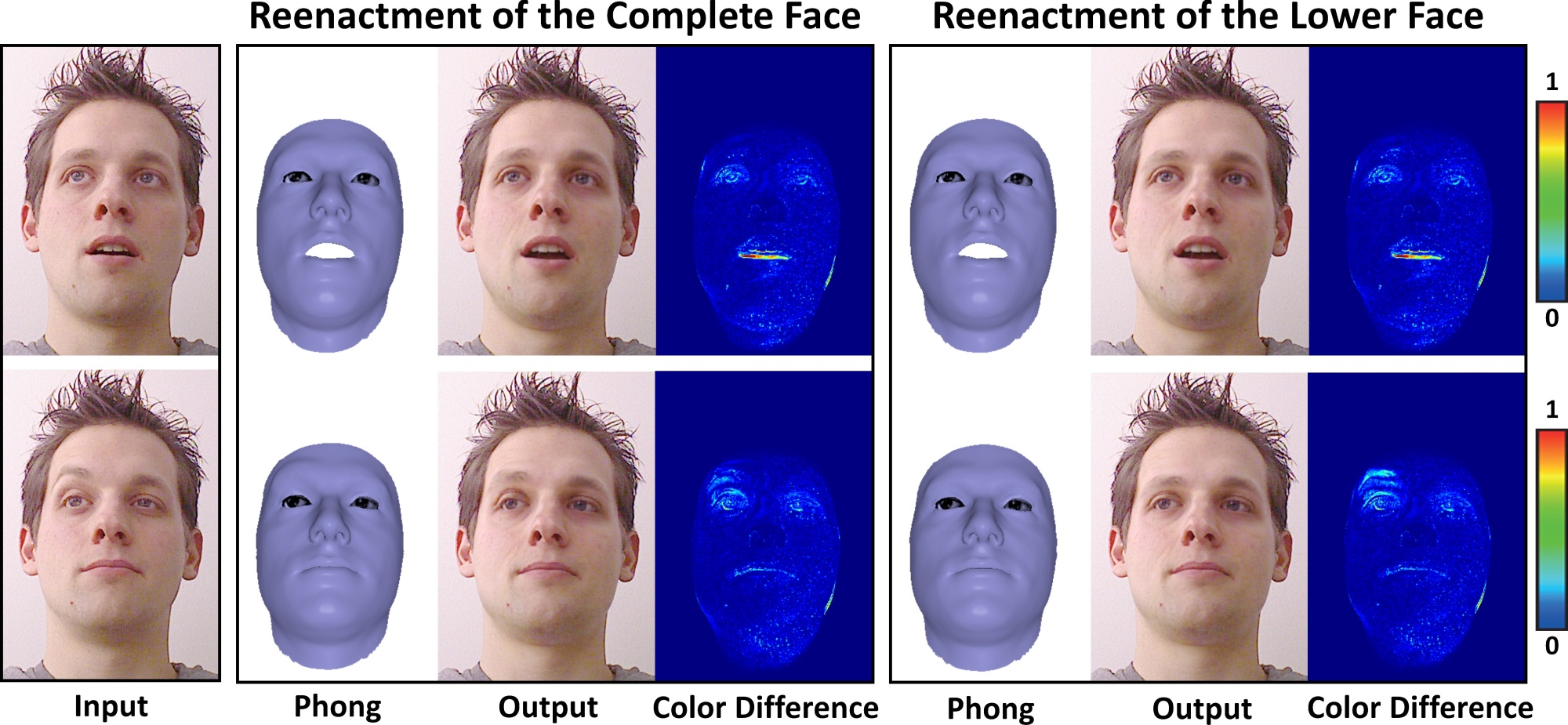}
	\vspace{-0.7cm}
	\caption{Perceptual side-by-side comparison for the self-reenactment scenario.}	
	\vspace{-0.5cm}
	\label{fig:full_vs_half_face_reenactment}
\end{figure}

To quantify the quality of our approach, we perform a side-by-side ground truth comparison for the self-reenactment scenario, see Fig.~\ref{fig:full_vs_half_face_reenactment}.
To this end, we employ the same sequence as source and as target.
This enables us to measure the color difference between the real video and the synthesized output.
In the VR scenario, the source is wearing an HMD, thus we are only able to track and transfer the expressions of the lower part of the face.
To measure the loss of information, we evaluate both scenarios, full reenactment and reenactment of only the lower part of the face.
We refer to the supplemental video for the complete video sequence.
Full facial reenactment results in a mean error of $0.01067$ measured in RGB color space.
Due to the lack of eyebrow motion, the reenactment of only the lower part of the face has a slightly higher error of $0.01086$.

We also conducted a pilot study with $18$ participants (working in the field of computer graphics) to evaluate the realism of our results.
A variety of different stereoscopic videos were shown.
The first video is a real video of an actor wearing an HMD, followed by result videos of our approach.
The participants were asked to rate the realism and the impression of sitting face-to-face to a person (from $1$ (very good) to $6$ (very bad)).
The original video achieved a score of $1.75$ and a score of $2.5625$, respectively.
The videos created with our stereoscopic reenactment method achieved a score of $2.281$ and $2.09$.
Our approach produces good quality and the preliminary perceptual evaluation shows that we improved the impression of sitting face-to-face to a person, which is of paramount importance for making VR teleconferencing viable.

%% file: limitations.tex

\begin{figure}
	\centering
	\includegraphics[width=0.85\linewidth]{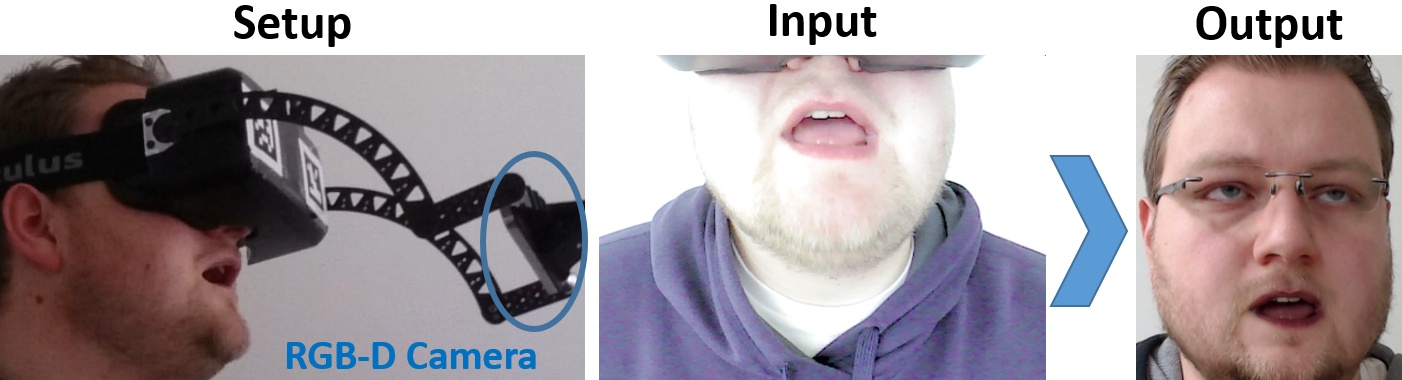}
	\vspace{-0.3cm}
	\caption{A setup with a rigidly mounted RGB-D camera (Intel Realsense F200) allows for cross-projection of the mouth independently of the head rotation.}
	\vspace{-0.4cm}
	\label{fig:new_setup}
\end{figure}

Although {\em FaceVR} is able to facilitate a wide range of face appearance manipulations in VR, it is one of the early methods in a new field.
As such, it is a first step and thus constrained by several limitations.
While our eye tracking solution provides great accuracy with little compute cost, it is specifically designed for the VR scenario. 
In contrast to \cite{Wang:2016} our approach is person-specific, but the \rev{person-specificity} allows us to re-synthesis eye motion photo-realistically.
Since our eye tracking approach is only based on one eye in the VR device, we \rev{cannot} correctly capture vergence and squinting;
one would need to add a second IR camera to the head mounted display, which is a straightforward modification.
As discussed in Sec.~\ref{sec:compositing}, we only employ one class for lid closure and apply a simple blending between open and closed eyes, explicitly modeling inbetween states can further improve the results \cite{Bermano:2015}.
The cross-projection of the mouth interior, which is used in the self-reenactment scenario, requires a similar head rotation in the source and target sequence.
If the head rotations differ too much, noticeable distortions might occur in the final output.
Therefore, we also tested a setup similar to Li et al.~\shortcite{li2015facial}, where the camera is rigidly attached to the HMD (see Fig.~\ref{fig:new_setup}).
Note that the original system of Li et al. is only able to animate a digital avatar and it does not allow for photo-realistic gaze-aware self-reenactment of a person.
The setup decreases the ergonomics of the HMD, but ensures a frontal view of the mouth that can be easily transfered to a front facing virtual stereoscopic avatar.
The major limitation of our approach is that we cannot modify the rigid head pose of the target videos. 
This would require a reconstruction of the background and the upper body of the actor including hair etc., which we believe is an interesting research direction.
Our VR face tracking is based on the rigid head pose estimates and the unoccluded face regions.
Unfortunately, the field of view of the IR camera attached to the inside of the device is not large enough to cover the entire occluded face region.
Thus, we cannot track most of the upper face except the eyeballs.
Here, our method is complementary to the approach of Li et al.~\shortcite{li2015facial}; they use additional sensor input from electronic strain measurements to fill in this missing data.
The resulting constraints could be easily included in our face tracking objective; note however, that their approach does not enable gaze-aware facial reenactment.
In the context of facial reenactment, we have similar limitations as Thies et al.~\shortcite{Thies15} and Face2Face~\cite{thies2016face}; i.e., we cannot handle occlusions in the target video such as those caused by microphones or waving hands.
We believe that this could be addressed by computing an explicit foreground-face segmentation; the work by Saito et al.~\shortcite{saito2016realtime} already shows promising results to specifically detect such cases.

%% file: conclusion.tex

In this work, we have presented {\em FaceVR}, a novel approach for real-time gaze-aware facial reenactment in the context of virtual reality.
The key components of FaceVR are robust face reconstruction and tracking, data-driven eye tracking, and photo-realistic re-rendering of facial content on stereo displays.
Therefore, we are able to show a variety of exciting applications, especially, self-reenactment for teleconferencing in VR.
We believe that this work is a stepping stone in this new field, demonstrating some of the possibilities of the upcoming virtual reality technology.
In addition, we are convinced that this is not the end of the line, and we believe that there will be even more exciting future work targeting photo-realistic video editing in order to improve the VR experience, as well as many other related applications.

%% file: appendix.tex

\begin{appendix}

\section{Appendix}
\label{sec:appendix}

In this appendix we show additional use-cases of {\em FaceVR}.
Beside self-reenactment for video conferences in VR, {\em FaceVR} produces compelling results for a variety of other applications, such as gaze-aware facial reenactment, reenactment in virtual reality, and re-targeting of somebody's gaze direction in a video conferencing call.

\subsection{Gaze-aware Facial Reenactment} \label{sec:gaze_reenactment}

\begin{figure}[h!]
	\centering
	\includegraphics[width=\linewidth]{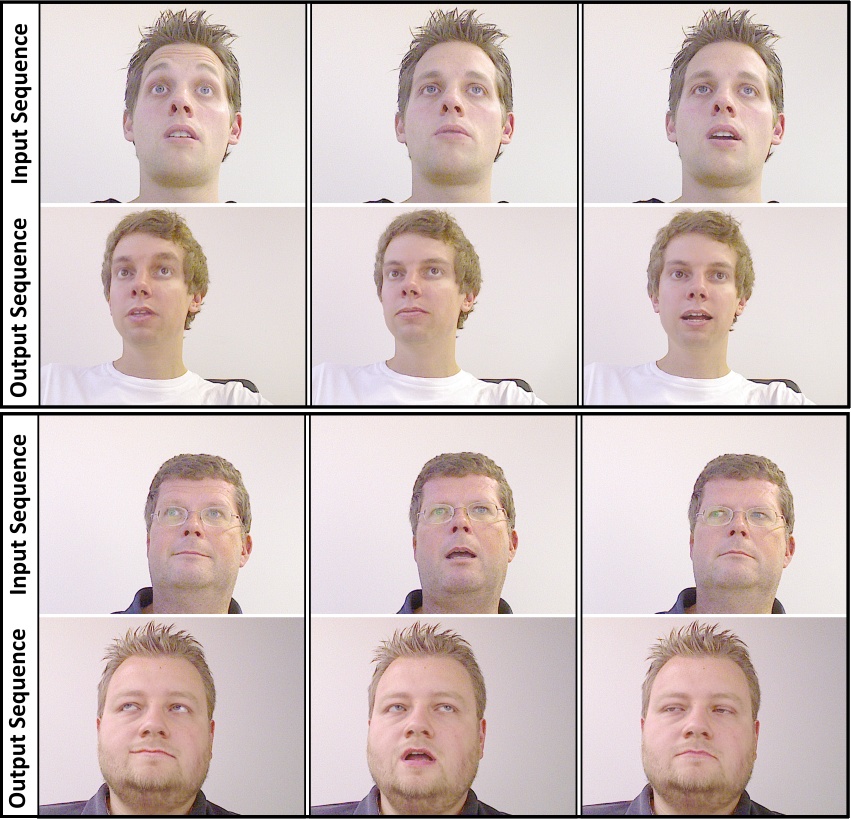}
	\caption{Gaze-aware facial reenactment of monocular RGB-D video streams: we employ our real-time performance capture and eye tracking approach in order to modify the facial expressions and eye motion of a target video. In each sequence, the source actor's performance (top) is used to drive the animation of the corresponding target video (bottom). Note, these results employ the mouth retrieval strategy to fill in the mouth interior.}
	\vspace{-0.2cm}
	\label{fig:mono_reenactment}
\end{figure}

\begin{figure}[h!]
	\centering
	\includegraphics[width=\linewidth]{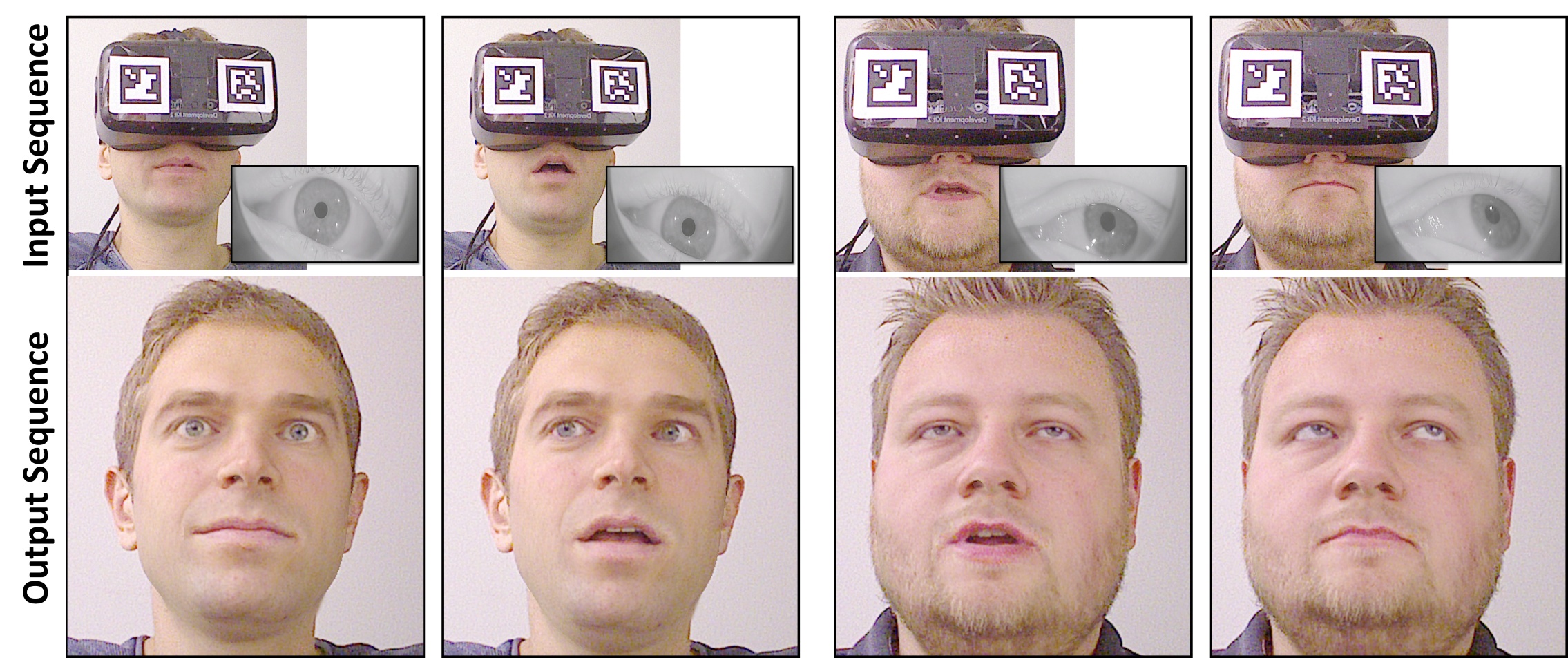}
	\caption{Self-Reenactment for VR Video Conferences: our real-time facial reenactment approach allows to virtually remove the HMD by driving a pre-recorded target video of the same person. Note, these results employ the mouth retrieval strategy to fill in the mouth interior}
	\vspace{-0.2cm}
	\label{fig:hmd_removal}
\end{figure}

\begin{figure}[h!]
	\centering
	\includegraphics[width=\linewidth]{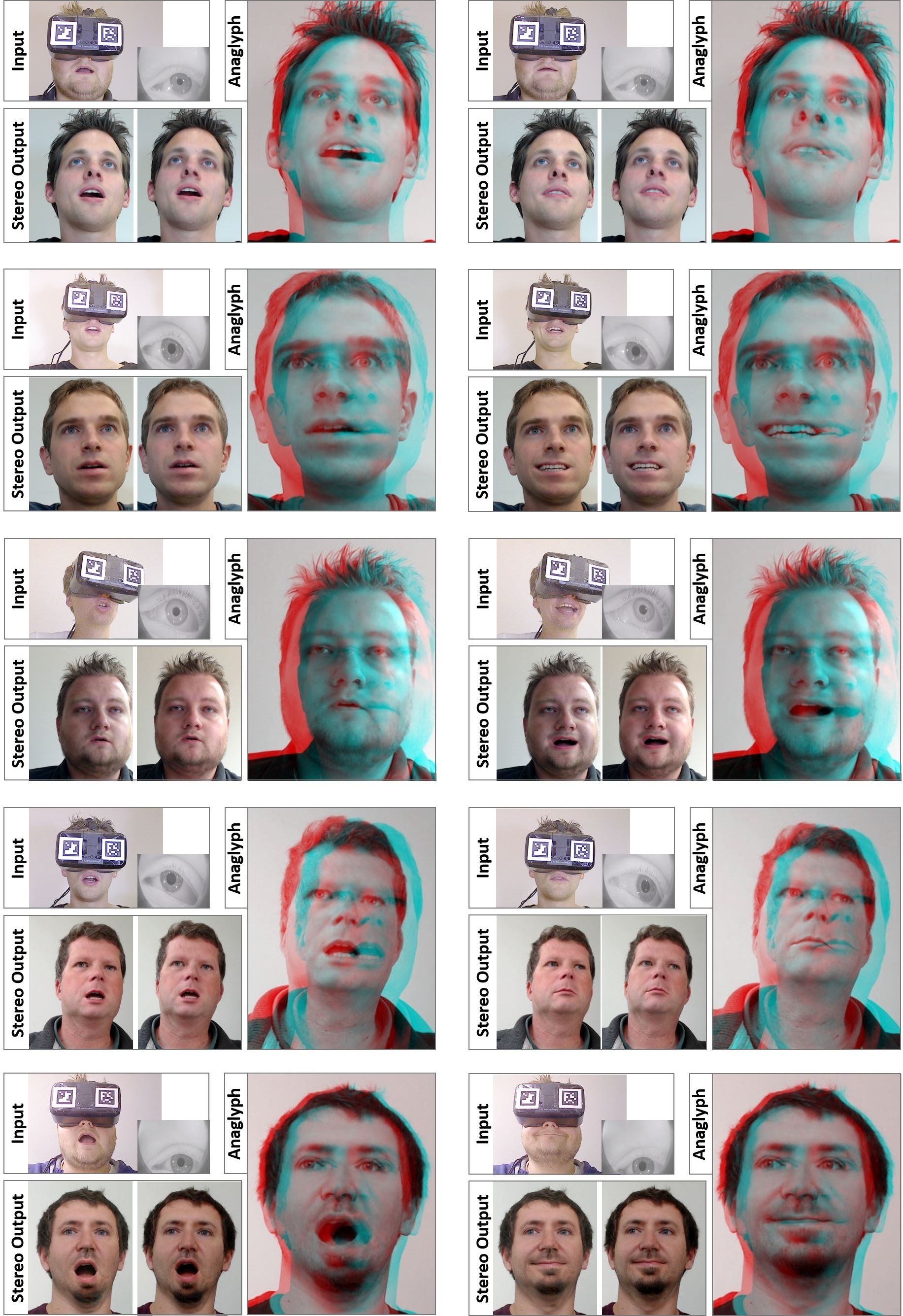}
	\caption{Gaze-aware facial reenactment of stereo target video content. We employ our real-time gaze-aware facial reenactment approach to modify the facial expressions and eye motion of stereo 3D content. The input (i.e., source actor) is captured with a frontal view and an internal IR camera. With our method, we can drive the facial animation of the stereo output videos shown below the input -- the facial regions in these images are synthetically generated. We employ the mouth retrieval strategy to fill in the mouth interior. The final results are visualized as anaglyph images on the right.}
	\vspace{-0.2cm}
	\label{fig:stereo_reenactment}
\end{figure}

Our approach enables real-time photo-realistic and gaze-aware facial reenactment of monocular RGB-D and 3D stereo videos; see Fig.~\ref{fig:stereo_reenactment}, \ref{fig:mono_reenactment} and \ref{fig:hmd_removal}.

\begin{figure}[b]
	\centering
	\includegraphics[width=0.8\linewidth]{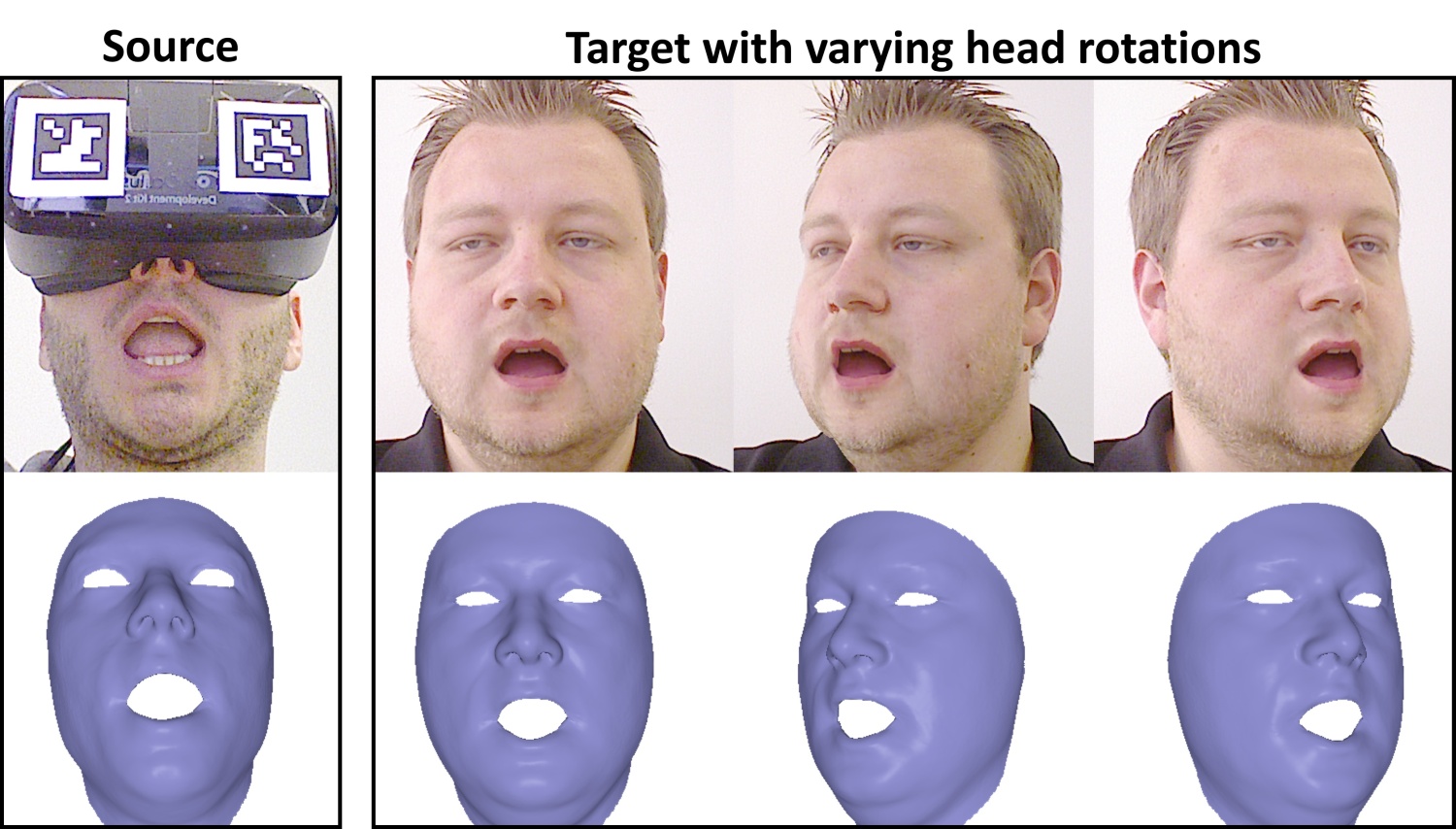}
	\caption{Reenactment results for different rigid head poses of the target actor. The mouth interior in the frontal view is of highest quality, since the mouth database consists of front facing mouth textures. Rigid rotations of the target actor's face still lead to plausible results with only minor distortions.}
	\label{fig:reenactment_head_rotation}
\end{figure}

In both scenarios, we track the facial expressions of a source actor using an external Asus Xtion RGB-D sensor, and transfer the facial expressions -- including eye motion -- to the video stream of a target actor.
The eye motion is tracked using our eye-gaze classifier based on the data captured by the external camera (monocular RGB-D reenactment) or the internal IR camera which is integrated into the HMD (stereo reenactment).
We transfer the tracked facial motion to a RGB-D or stereo target video stream using the presented facial reenactment approach.
The modified eye region is synthesized using our unified image-based eye and eyelid model (see main paper for more details).
This allows the source actor to take full control of the face expression and eye gaze of the target video stream at real-time frame rates.
Our approach leads to plausible reenactment results even for greatly differing head poses in the target video, see Fig.~\ref{fig:reenactment_head_rotation}.

\subsection{Gaze Correction for Video Conferencing} \label{sec:gaze_correction}
\begin{figure}[b]
	\centering
	\includegraphics[width=0.8\linewidth]{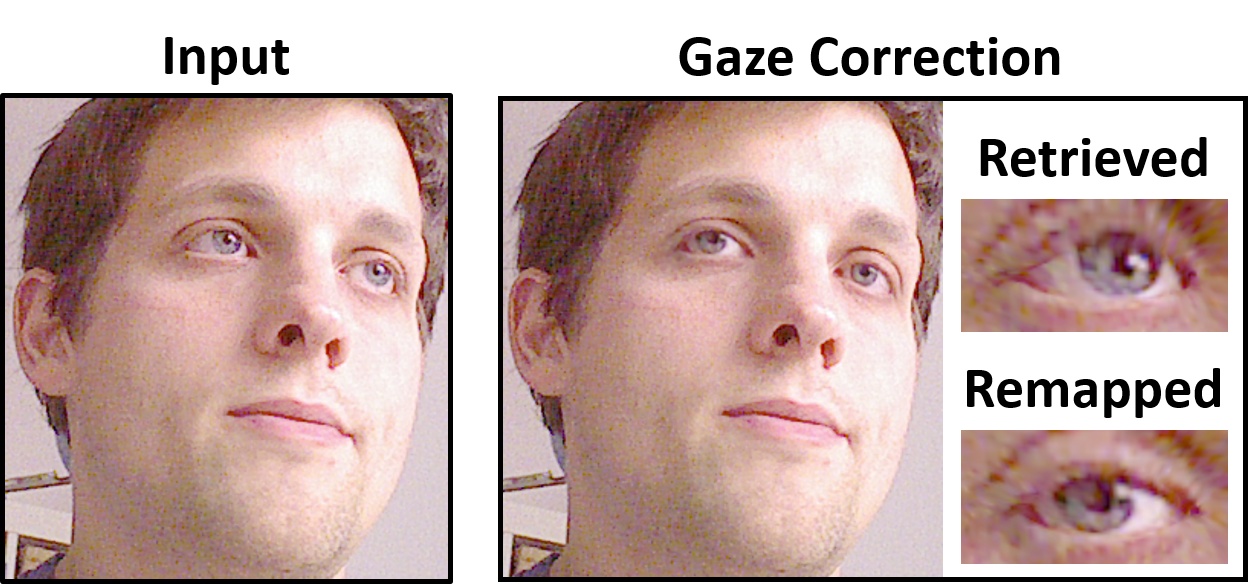}
	\caption{Gaze Correction: a common problem in video chats is the discrepancy between the physical location of the webcam and the screen, which leads to unnatural eye appearance (left). We use our eye tracking and retrieval strategy to correct the gaze direction in such a scenario, thus enabling realistic video conversations with natural eye contact (right).
	}
	\label{fig:video_conference}
\end{figure}

Video conference calls, such as Skype chats, suffer from a lack of eye contact between participants due to the discrepancy between the physical location of the camera and the screen.
To address this common problem, we apply our face tracking and reenactment approach to the task of online gaze correction for monocular live video footage; see Fig.~\ref{fig:video_conference}.
Our goal is the photo-realistic modification of the eye motion in the input video stream using our image-based eye and eyelid model.
To this end, we densely track the face of the user, and our eye-gaze classifier provides us with an estimate of the gaze direction; i.e., we determine the 2D screen position where the user is currently looking.
Given the eye tracking result, we modify the look-at point by applying a delta offset to the gaze direction which corrects for the different positions of the camera and screen.
Finally, we retrieve a suitable eye texture that matches the new look-at point and composite it with the monocular input video stream to produce the final output.
A gaze correction example is shown in Fig.~\ref{fig:video_conference}.

\end{appendix}